\documentclass{article}

\usepackage{config}

\usepackage[preprint]{neurips_2024}

\usepackage[utf8]{inputenc} % allow utf-8 input
\usepackage[T1]{fontenc}    % use 8-bit T1 fonts
\usepackage{hyperref}       % hyperlinks
\usepackage{url}            % simple URL typesetting
\usepackage{booktabs}       % professional-quality tables
\usepackage{amsfonts}       % blackboard math symbols
\usepackage{nicefrac}       % compact symbols for 1/2, etc.
\usepackage{microtype}      % microtypography
\usepackage{xcolor}         % colors

\hypersetup{
    colorlinks=true,
    linkcolor=blue, % Modify these colors as per your preference
    filecolor=blue,      
    urlcolor=blue,
    citecolor=blue
}

\title{Toward Global Convergence of  Gradient EM for Over-Parameterized Gaussian Mixture Models}

\author{%
  Weihang Xu%\thanks{Use footnote for providing further information about author (webpage, alternative address)---\emph{not} for acknowledging funding agencies.} 
    \\
 University of Washington\\
  \texttt{xuwh@cs.washington.edu}
  \And
 Maryam Fazel\\
  University of Washington\\
  \texttt{mfazel@uw.edu}
  \And
  Simon S. Du\\
   University of Washington\\
  \texttt{ssdu@cs.washington.edu}
}

\begin{document}

\maketitle

\begin{abstract}
We study the gradient Expectation-Maximization (EM) algorithm for Gaussian Mixture Models (GMM) in the over-parameterized setting, where a general GMM with $n>1$ components learns from data that are generated by a single ground truth Gaussian distribution. 
While results for the special case of 2-Gaussian mixtures are well-known, a general global convergence analysis for arbitrary $n$ remains unresolved and faces several new technical barriers since the convergence becomes sub-linear and non-monotonic.
To address these challenges, we construct a novel likelihood-based convergence analysis framework and rigorously prove that gradient EM converges globally with a sublinear rate $O(1/\sqrt{t})$. This is the first global convergence result for Gaussian mixtures with more than $2$ components.
The sublinear convergence rate is due to the algorithmic nature of learning over-parameterized GMM with gradient EM. We also identify a new emerging technical challenge for learning general over-parameterized GMM: the existence of bad local regions that can trap gradient EM for an exponential number of steps.

%providing the first global convergence proof of gradient EM beyond the special case of $n=2$. 
% We further present a matching  $\Omega(1/\sqrt{T})$ convergence rate lower bound, implying the tightness of our convergence bound and revealing a slow-down effect on convergence caused by over-parameterization.
%This result shows that gradient EM converges exponentially slower in the over-parameterized setting, compared to the exact parameterized setting where it enjoys a $\exp(-\Omega(T))$ linear convergence rate. 
% Our proof of the convergence rate lower bound is based on another novel potential function and explains that the cause of the slower convergence rate is %due to 
% additional degrees of freedom in the parameter space introduced by over-parameterization.  %\mf{maybe remove the rest of this sentence, might sound too simplistic:}
\end{abstract}

\section{Introduction}\label{Section intro}
Learning Gaussian Mixture Models (GMM) is a fundamental problem in machine learning with broad applications. In this problem, data generated from a mixture of $n\geq 2$ ground truth Gaussians are observed without the label (the index of component Gaussian that data is sampled from), and the goal is to retrieve the maximum likelihood estimation of Gaussian components. The Expectation Maximization (EM) algorithm is arguably the most widely-used algorithm for this problem. Each iteration of the EM algorithm consists of two steps. In the expectation (E) step, it computes the posterior probability of unobserved mixture membership label according to the current parameterized model. In the maximization (M) step, it computes the maximizer of the $Q$ function, which is the likelihood with respect to posterior estimation of the hidden label computed in the E step. 

Gradient EM, as a popular variant of EM, is often used in practice when the maximization step of EM is costly or even intractable. It replaces the M step of EM with taking one gradient step on the $Q$ function. Learning Gaussian Mixture Models with EM/gradient EM is an important and widely-studied problem. Starting from the seminal work \citep{balakrishnan2014statistical}, a flurry of work \cite{daskalakis17TenSteps,Xu2016GlobalAO,Dwivedi2018TheoreticalGF,kwon_em_2020,Dwivedi2019SharpAO} have studied the convergence guarantee for EM/gradient EM in various settings. However, these works either only prove local convergence, or consider the special case of $2$-Gaussian mixtures. A general global convergence analysis of EM/gradient EM on $n$-Gaussian mixtures still remains unresolved. \citet{Jin2016LocalMI} is a notable negative result in this regard, where the authors show that on GMM with $n\geq 3$ components, randomly initialized EM will get trapped in a spurious local minimum with high probability. 

\bftext{Over-parameterized Gaussian Mixture Models.} 
Motivated by the negative results, a line of work considers the over-parameterized setting where the model uses more Gaussian components than the ground truth GMM, in the hope that it might help the global convergence of EM and bypass the negative result. 
%This over-parameterized regime is also more reasonable since it does not assume prior knowledge on the number of components in ground truth GMM.
In such over-parameterized regime, the best that people know so far is from \citep{Dwivedi2018SingularityMA}. This work proves global convergence of 2-Gaussian mixtures on one single Gaussian ground truth. The authors also show that EM has a unique sub-linear convergence rate in this over-parameterized setting (compared with the linear convergence rate in the exact-parameterized setting \citep{balakrishnan2014statistical}). This motivates the following natural open question:
\begin{center}
    \emph{Can we prove global convergence of the EM/gradient EM algorithm on general $n$-Gaussian mixtures in the over-parameterized regime?}
\end{center}

In this paper, we take a significant step towards answering this question. Our main contributions can be summarized as follows:
\begin{itemize}
    \item We prove global convergence  of the gradient EM algorithm for learning general $n$-component GMM on one single ground truth Gaussian distribution. This is, to the best of our knowledge, the first global convergence proof for general $n$-component GMM. Our convergence rate is sub-linear, reflecting an inherent nature of over-parameterized GMM (see Remark \ref{Remark sublinear} for details).
    
    \item We propose a new analysis framework that utilizes the likelihood function for proving convergence of gradient EM.
    Our new framework tackles several emerging technical barriers for global analysis of general GMM.
    \item We also identify a new geometric property of gradient EM for learning general $n$-component GMM: There exists bad initialization regions that traps gradient EM for exponentially long, resulting in an inevitable exponential factor in the convergence rate of gradient EM.
\end{itemize}

% Outline for intro
% \begin{itemize}
%     \item 1st paragraph, general introduction to EM and GMM and gradient EM (adapt prior paper).
%     \item What did prior theoretical work do (overview): either local analysis, global convergence negative result from Chi Jin,  2 to 2 convergence, 2 to 1 with symmetric mean (mention they got sublinear convergence). More details in Related Work section xxx.  Open problem: global convergence for multiple Gaussian, especially in the over-parameterized regime.
%     \item Subsection of Main contributions: we take an important step: on n to 1 regime. Give an informal theorem. We tackle several technical barriers. Also mention theorem 7.
% \item subsection of technical overview: summarize Section 4 (each subsection in Section 4 corresponds to one paragraph).
% \end{itemize}

\subsection{Gaussian Mixture Model (GMM)}
We consider the canonical Gaussian Mixture Models with weights $\bm{\pi}=(\pi_1,\ldots, \pi_n)$ ($\sum_{i=1}^n \pi_i=1$), means $\bm{\mu} =(\mu_1^{\top}, \ldots, \mu_n^{\top})^{\top}$ and unit covariance matrices $I_d$ in $d$-dimensional space. Following a widely-studied setting \citep{balakrishnan2014statistical, yan_convergence_2017, daskalakis17TenSteps}, we set the weights $\bm{\pi}$ and covariances $I_d$ in student GMM as fixed, and the means $\bm{\mu} =(\mu_1^{\top}, \ldots, \mu_n^{\top})^{\top}$ as trainable parameters. We use $\GMM(\bmu)$ to denote the GMM model parameterized by $\bm{\mu} $, which can be described with probability density function (PDF) $p_{\bmu}: \R^d\to \R_{\geq 0}$ as
\begin{equation}\label{GMM}
    p_{\bmu}(x) =\sum_{i\in[n]} \pi_i \phi(x|\mu_i, I_d) = \sum_{i\in[n]} \pi_i (2\pi)^{-d/2}\exp\left(-\frac{\|x-\mu_i\|^2}{2}\right),
\end{equation}
where $\phi(\cdot|\mu, \Sigma)$ is the PDF of $\N(\mu, \Sigma)$, $\pi_1+\cdots+\pi_n=1, \pi_i>0, \forall i\in[n]$.

\subsection{Gradient EM algorithm}
The EM algorithm is one of the most popular algorithms for retrieving the maximum likelihood estimator (MLE) on latent variable models. In general, EM and gradient EM address the following problem: given a joint distribution $p_{\bmu^*}(x,y)$ of random variables $x,y$ parameterized by $\bmu^*$, observing only the distribution of $x$, but not the latent variable $y$, the goal of EM and gradient EM is to retrieve the maximum likelihood estimator
\[\hat{\bmu}_{\text{MLE}} \in \arg\max_{\bmu}\log p_{\bmu}(x).\]
The focus of this paper is the non-convex optimization analysis, so we consider using \emph{population gradient EM} algorithm to learn GMM \eqref{GMM}, where the observed variable is $x\in \R^d$ and latent variable is the index of membership Gaussian in GMM. We follow the standard teacher-student setting where a student model $\GMM(\bmu)$ with $n\geq 2$ Gaussian components learns from data generated from a ground truth teacher model $\GMM(\bmu^*)$. 
We consider the over-parameterized setting where the ground truth model $\GMM(\bmu^*)$ is a single Gaussian distribution $\N(\mu^*, I_d)$, namely $ \bm{\mu}^* =({\mu^*}^{\top}, \ldots, {\mu^*}^{\top})^{\top}$. We can then further assume $w.l.o.g.$ that $\mu^*=0$. Our problem could be seen as a strict generalization of \cite{Dwivedi2018SingularityMA}, where they studied using mixture model of \emph{two Gaussians} with symmetric means (they set constraint $\mu_2 = -\mu_1$) to learn one single Gaussian.

At time step $t=0,1,2,\ldots$, given with parameters $\bm{\mu}(t) = (\mu_1(t)^{\top},\ldots, \mu_n(t)^{\top})^{\top}$, population gradient EM updates $\bm{\mu}$ via the following two steps
\begin{itemize}
    \item E step:  for each $i\in[n]$, compute the membership weight function $\psi_i: \R^d\to \R$ defined as
    \begin{equation}\label{psi i definition}
        \psi_i(x|\bmu(t)) =\Pr [i|x]=\frac{\pi_i\exp\left(-\frac{\|x-\mu_i(t)\|^2}{2}\right)}{\sum_{k\in[n]} \pi_k\exp\left(-\frac{\|x-\mu_k(t)\|^2}{2}\right)}.
    \end{equation}

    \item M step: Define $Q(\cdot|,\mu(t))$ as 
    \[Q(\bm{\mu} | \bm{\mu}(t))=\E_{x\sim \N(0,I_d)} \left[\sum_{i=1}^n -\psi_i(x| \bmu(t))\frac{\|x-\mu_i\|^2}{2}\right],\]
    
    Gradient EM with step size $\eta>0$ performs the following update:
    \begin{equation}\label{gradient EM update}
        \mu_i(t+1) = \mu_i(t) + \eta \nabla_{\mu_i} Q(\bm{\mu}(t)|\bm{\mu}(t))= \mu_i(t)-\eta\E_{x\sim \N(0,I_d)} \left[\-\psi_i(x|\bmu(t))(\mu_i(t)-x)\right].
    \end{equation}
    % In this paper, we consider gradient EM flow, $i.e.,$ gradient EM with infinitesimal step size. When the step size $\eta\to 0$, the dynamics of $\bmu$ is described by the following ODE
    % \[\frac{\partial}{\partial t}\mu_i(t) = -\nabla_{\mu_i} Q(\bm{\mu}(t)|\bm{\mu}(t))=-\E_{x\sim \N(0,I_d)} \left[\-\psi_i(x|\bmu(t))(\mu_i(t)-x)\right]. \forall t    .\]
\end{itemize}

The membership weight function $x\to \psi_i(x|\bmu)$ represents the posterior probability of data point $x$ being sampled from the $i^{\text{th}}$ Gaussian of $\GMM(\bmu)$.
For ease of notation, we sometimes simply write $\psi_i(x|\bmu)$ as $\psi_i(x)$ when the choice of $\bmu$ is obvious.

\subsection{Loss function of gradient EM}\label{gradient EM = GD}
Since the task of gradient EM is to find the MLE over ground truth distribution $p_{\bmu^*}$, we can define the MLE loss function for gradient EM as 
\begin{equation}\label{loss}
    \loss(\bmu) = \kl(p_{\bmu^*}||p_{\bmu})=-\E_{x\sim p_{\bmu^*}}\left[\log\left(\frac{p_{\bmu}(x)}{p_{\bmu^*}(x)}\right)\right].
\end{equation}
The loss $\loss$ is the Kullback–Leibler (KL) divergence between the ground truth GMM and the student model GMM. Since finding MLE is equivalent to minimizing the KL divergence between model and the ground truth, the goal of gradient EM is equivalent to finding the global minimum of loss $\loss$. In other words, proving that gradient EM finds the MLE is equivalent with proving the convergence of $\loss$ to $0$. However, we are going to present another reason why loss function $\loss$ is important, for it is also closely related to the dynamics of gradient EM.

\bftext{Gradient EM is gradient descent on $\loss$.} We present the following important observation. The proof is deferred to appendix.
\begin{fact}\label{Fact}
    For any $\bmu$, $ \nabla Q(\bmu|\bmu)=-\nabla \loss(\bmu)$.
\end{fact}

% \simon{put proof in the appendix.}

Fact \ref{Fact} states that the gradient of $Q$ function that gradient EM optimizes in each iteration is identical to the (inverse of) gradient of loss function $\loss$. This observation is very useful since it implies that gradient EM is equivalent to gradient descent (GD) algorithm on $\loss$. This observation is not a new discovery of ours but actually a wide-spread folklore (see \citep{Jin2016LocalMI}). However, our new contribution is %that we found 
to observe Fact \ref{Fact} is very helpful for analyzing gradient EM, and to construct a new convergence analysis framework for gradient EM based on it.

\subsection{Notation}
In this paper, we adopt the following  notational conventions. We denote $\{1,2,\ldots,n\}$ with $[n]$. $\bm{\mu} =(\mu_1^{\top}, \ldots, \mu_n^{\top})^{\top}\in\R^{nd}$ denotes the parameter vector of GMM obtained by concatenating Gaussian mean vectors $\mu_1, \ldots, \mu_n$ together. For any vector $\mu$, $\mu(t)$ denotes its value at time step $t$, sometimes we omit this iteration number $t$ when its choice is clear and simply abbreviate $\mu(t)$ as $\mu$. We define a shorthand of expectation taken over the ground truth GMM $\E_{x\sim\N(0, I_d)}[\cdot]$ as $\E_x[\cdot]$. For any vector $v\neq 0$, we use $\overline{v}\coloneqq v/\|v\|$ to denote the normalization of $v$. We define (with a slight abuse of notation) $\imax \coloneqq \arg\max_{i\in [n]}\{\|\mu_i\|\}$ as the index of  $\mu_i$ with the maximum norm, and $\mumax \coloneqq \|\mu_{\imax}\|=\max_{i\in [n]}\{\|\mu_i\|\}$ as the maximum norm of $\mu_i$. In particular, $\mumax(t)=\max\{\|\mu_1(t)\|,\ldots, \|\mu_n(t)\|\}$. Similarly, $\pimin\coloneqq \min_{i\in[n]}\pi_i$ and $\pimax\coloneqq \max_{i\in[n]}\pi_i$ denotes the minimal and maximal $\pi_i$, respectively. We use $\nabla_{\mu_i}\loss$ to denote the gradient of $\mu_i$ on $\loss$, and $\nabla \loss =(\nabla_{\mu_1}\loss^{\top},\ldots,\nabla_{\mu_n}\loss)^{\top}$ denotes the collection of all gradients. Finally we define a potential function $U:\R^{nd}\to \R \;$ for $\GMM(\bmu)$ as
    \[U(\bmu)=\sum_{i\in[n]}\|\mu_i\|^2.\]

\subsection{Technical overview}
Here we provide a brief summary of the major technical barriers for our global convergence analysis and our techniques for overcoming them.

\bftext{New likelihood-based analysis framework.} The traditional convergence analysis for EM/gradient EM in previous works \cite{balakrishnan2014statistical, yan_convergence_2017, kwon_em_2020} proceeds by showing the distance between the model and the ground truth GMM in the \emph{parameter space} contracts linearly in every iteration. This type of approach meets new challenges in the over-parameterized $n$-Gaussian mixture setting since the convergence is both sub-linear and non-monotonic. To address these problems, we propose a new likelihood-based convergence analysis framework: instead of proving the convergence of parameters, our analysis proceeds by showing the likelihood loss function $\loss$ converges to $0$. The new analysis framework is more flexible and allows us to overcome the aforementioned technical barriers.

\bftext{Gradient lower bound. }The first step of our global convergence analysis constructs a gradient lower bound. Using some algebraic transformation techniques, we convert the gradient projection $\ip{\loss(\bmu),\bmu}$ into the expected norm square of a random vector $\tpsi(x)$. (See Section \eqref{Section proof overview} for the full definition). Although lower bounding the expectation of $\tpsi$ is very challenging, our key idea is that the gradient of $\tpsi$ has very nice properties and can be easily lower bounded, allowing us to establish the gradient lower bound.

\bftext{Local smoothness and regularity condition. } After obtaining the gradient lower bound, the missing component of the proof is a smoothness condition of the loss function $\loss$. Since proving the smoothness of $\loss$ is hard in general, we define and prove a weaker notion of local smoothness, which suffices to prove our result. In addition, we design and use an auxiliary function $U$ to show that gradient EM trajectory satisfies the locality required by our smoothness lemma.

\section{Related work}\label{Section related works}

\subsection{2-Gaussian mixtures}
There is a vast literature studying the convergence of EM/gradient EM on $2$-component GMM. The initial batch of results proves convergence within a infinitesimally small local region \citep{xu1996convergence, ma2000asymptotic}.
\citet{balakrishnan2014statistical} proves for the first time convergence of EM and gradient EM within a non-infinitesimal local region. Among the %followed 
later works on the same problem, \citet{Klusowski2016StatisticalGF} improves the basin of convergence guarantee, \citet{daskalakis17TenSteps,Xu2016GlobalAO} proves the global convergence for $2$-Gaussian mixtures. 
These works focused on the exact-parameterization scenario where the number of student mixtures is the same as that of the ground truth.
More recently, \citet{wu2019randomly} proves global convergence of $2$-component GMM without any separation condition. 
Their result can be viewed as a convergence result in the over-parameterized setting where the student model has two Gaussians and the ground truth is a single Gaussian.
On the other hand, their setting is more restricted than ours because they require the means of two Gaussians in the student model to be symmetric around the ground truth mean.
\citet{Weinberger2021TheEA} extends the convergence guarantee to the case of unbalanced weights.
Another line of work \cite{Dwivedi2018SingularityMA, Dwivedi2019SharpAO,Dwivedi2018TheoreticalGF} studies the over-parameterized setting of using $2$-Gaussian mixture to learn a single Gaussian and proves global convergence of EM. Our result extends this type of analysis to the general case of $n$-Gaussian mixtures, which requires significantly different techniques.
We note that going beyond Gaussian mixture models, there are also works studying EM algorithms for other mixture models such as a mixture of linear regression~\cite{kwon2019global}.

\subsection{N-Gaussian mixtures}
Another line of results focuses on the general case of $n$ Gaussian mixtures. \citet{Jin2016LocalMI} provides a counter-example showing that EM does not converge globally for $n>2$ (in the exact-parameterized case). 
\citet{dasgupta2013two} prove that a variant of EM converges to MLE in two rounds for $n$-GMM. Their result relies on a modification of the EM algorithm and is not comparable with ours.
\citep{chen2023local} analyzes the structure of local minima in the likelihood function of GMM. However, their result is purely geometric and does not provide any convergence guarantee.

A series of paper \cite{yan_convergence_2017, 
 zhao_statistical_2018,kwon_em_2020, segol_improved_2021} follow the framework proposed by \cite{balakrishnan2014statistical} to prove the \emph{local} convergence of EM for $n$-GMM. While their result applies to the more general $n$-Gaussian mixture ground truth setting, their framework only provides local convergence guarantee and cannot be directly applied to our setting.

 % \mf{here's the other Caramanis paper on EM for mixed linear regression: https://proceedings.mlr.press/v130/kwon21b/kwon21b.pdf}

\subsection{Slowdown due to over-parameterization} This paper gives an $O\left(1/\sqrt{t}\right)$ bound %of 
for fitting over-parameterized Gaussian mixture models to a single Gaussian. Recall that to learn a single Gaussian, if one's student model is also a single Gaussian, then one can obtain an $\exp(-\Omega(t))$ rate because the loss is strongly convex. This slowdown effect due to over-parameterization has been observed for Gaussian mixtures in \cite{Dwivedi2018TheoreticalGF,wu2019randomly}, but has also been observed in other learning problems, such as learning a two-layer neural network~\cite{xu2023over,richert2022soft} and matrix sensing problems~\citep{xiong2023over,zhang2021preconditioned,zhuo2021computational}.
\section{Main results}\label{Section Main Result}
In this section, we present our main theoretical result, which consists of two parts: In Section \ref{Section global convergence} we present our global convergence analysis of gradient EM, in Section \ref{Section counter-examples} we prove that an exponentially small factor in our convergence bound is inevitable and cannot be removed. All omitted proofs are deferred to the appendix.

\subsection{Global convergence of gradient EM}
\label{Section global convergence}
We first present our main result, which states that gradient EM converges to MLE globally.

\begin{theorem}[Main result]\label{Global convergence of gradient EM}
Consider training a student $n$-component GMM initialized from $\bm{\mu}(0) = (\mu_1(0)^{\top},\ldots, \mu_n(0)^{\top})^{\top}$ to learn a single-component ground truth GMM $\N(0, I_d)$ with population gradient EM algorithm. If the step size satisfies $\eta \leq O\left(\frac{\exp\left(-8U(0)\right)\pimin^2}{n^2d^2(\frac{1}{\mumax(0)}+\mumax(0))^2}\right)$, then gradient EM converges globally with rate
\[\loss(\bmu(t))\leq \frac{1}{\sqrt{\gamma t}},\]
where  $\gamma = \Omega\left(\frac{\eta\exp\left(-16U(0)\right)\pimin^4}{n^2d^2(1+\mumax(0){\sqrt{dn}})^4}\right)\in\R^+$. Recall that $\mumax(0)=\max\{\|\mu_1(0)\|,\ldots, \|\mu_n(0)\|\}$ and $U(0)=\sum_{i\in[n]}\|\mu_i(0)\|^2$ are two initialization constants.

% If step size $\eta \leq O\left(\frac{\exp\left(-8n\mumax^2(0)\right)\pimin^2}{n^2d^2(\frac{1}{\mumax(0)}+\mumax(0))^2}\right)$, then population gradient EM algorithm on GMM initialized from point $\bm{\mu}(0) = (\mu_1(0)^{\top},\ldots, \mu_n(0)^{\top})^{\top}$ converges globally with rate
% \[\loss(\bmu(t))\leq \frac{1}{\sqrt{\gamma t}},\]
% where $\gamma = \Omega\left(\frac{\eta\exp\left(-16n\mumax^2(0)\right)\pimin^4}{n^2d^2(1+\mumax(0){\sqrt{dn}})^4}\right)\in\R^+$ and $\mumax(0)=\max\{\|\mu_1(0)\|,\ldots, \|\mu_n(0)\|\}$.
\end{theorem}

\begin{remark}\label{Remark sublinear}
Without over-parameterization, for learning a single Gaussian, one can obtain a linear convergence $\exp (-\Omega\left(t\right))$.
    We would like to note that the sub-linear convergence rate guarantee of gradient EM stated in Theorem \ref{Global convergence of gradient EM} ($\loss(\bmu(t))\leq O(1/\sqrt{t})$) is due to the inherent nature of the algorithm. \citet{Dwivedi2018SingularityMA} studied the special case of using 2 Gaussian mixtures with symmetric means to learn a single Gaussian and proved that EM has sublinear convergence rate when the weights $\pi_i$ are equal. Since Theorem \ref{Global convergence of gradient EM} studies the more general case of $n$ Gaussian mixtures, this type of subexponential convergence rate is the best than we can hope for. 
\end{remark}

\begin{remark}
    The convergence rate in Theorem \ref{Global convergence of gradient EM} has a factor exponentially small in the initialization scale ($\gamma \propto\exp(-16U(0))$). We would like to stress that this is again due to algorithmic nature of the problem rather than the limitation of analysis. In Section \ref{Section counter-examples}, we prove that there exists bad regions with exponentially small gradients so that when initialized from such region, gradient EM gets trapped locally for $\exp(\Omega(U(0)))$ number of steps. Therefore, a convergence speed guarantee exponentially small in $U(0)$ is inevitable and cannot be improved.
\end{remark}

\begin{remark}\label{parametric convergence}
    Theorem \ref{Global convergence of gradient EM} is fundamentally different from convergence analysis for EM/gradient EM in previous works \cite{yan_convergence_2017,Dwivedi2019SharpAO,balakrishnan2014statistical} which proved monotonic linear contraction of parameter distance $\|\bmu(t)-\bmu^*\|$. But our result also implies global convergence since loss function $\loss$ converging to $0$ is equivalent to convergence of gradient EM to MLE.
\end{remark}

\begin{remark} The convergence result in Theorem \ref{Global convergence of gradient EM} is for population gradient EM, but it also implies global convergence for sample-based gradient EM as the sample size tends to infinity. For a similar reduction from population EM to sample EM, see Section 2.2 of \citep{Xu2016GlobalAO}.
    
\end{remark}

\subsection{Necessity of exponentially small factor in convergence rate} \label{Section counter-examples}

In this section we prove that a factor exponentially small in initialization scale ($\exp(-\Theta(U(0)))$) is inevitable in the global convergence rate guarantee of gradient EM. Particularly, we show the existence of  bad regions such that %when initialized from, 
initialization from this region traps gradient EM for exponentially long time before final convergence. Our result is the following theorem.

\begin{theorem}[Existence of bad initialization region]\label{Existence of bad initialization region}
For any $n\geq 3$, define $\tilde{\bmu}(0)=({\mu}^{\top}_1(0),\ldots,{\mu}^{\top}_n(0))$ as follows: $ {\mu}_1(0)=12\sqrt{d}e_1, {\mu}_2(0)=-12\sqrt{d}e_1, \mu_3(0)=\cdots=\mu_n(0)=0$, where $e_1$ is a standard unit vector. Then population gradient EM initialized with means $\tilde{\bmu}(0)$ and equal weights $\pi_1=\ldots=\pi_n=1/n$ will be trapped in a bad local region around $\tilde{\bmu}(0)$ for exponentially long time \[T\coloneqq\frac{1}{30\eta}e^{d}=\frac{1}{30\eta}\exp(\Theta(U(0))).\] More rigorously, for any $0\leq t\leq T, \exists i\in[n]$ such that
    \[ \|\mu_i(t)\|\geq 10\sqrt{d}. \;\; \]
\end{theorem}

Theorem \ref{Existence of bad initialization region} states that, when initialized from some bad points $\bmu(0)$, after $\exp(\Theta(U(0)))$ number of time steps, gradient EM will still stay in this local region and remain $10\sqrt{d}$ distance away from the global minimum $\bmu=0$. Therefore an exponentially small factor in convergence rate is inevitable.

\begin{remark}
    Theorem \ref{Existence of bad initialization region} eliminates the possibility of proving any polynomial convergence rate of gradient EM from arbitrary initialization. However, it is still possible to prove that, with some specific smart initialization schemes, gradient EM avoids the bad regions stated in Theorem \ref{Existence of bad initialization region} and enjoys a polynomial convergence rate. We leave this as an interesting open question for future analysis.
\end{remark}
\section{Proof overview}\label{Section proof overview}
In this section, we provide a technical overview of the proof in our main result (Theorem \ref{Global convergence of gradient EM} and Theorem \ref{Existence of bad initialization region}).

% \subsection{Proof road map}

\subsection{Difficulties of a global convergence proof and our new analysis framework}\label{Section Hardness}

Proving the global convergence of gradient EM for general $n$-Gaussian mixture is highly nontrivial. While there have been many previous works \citep{balakrishnan2014statistical,yan_convergence_2017,Dwivedi2018SingularityMA} studying either local convergence or the special case of $2$-Gaussian mixtures, they all focus on showing the contraction of parametric error. Namely, their proof proceeds by showing the distance between the model parameter and the ground truth contracts, usually by a fixed linear ratio, in each iteration of the algorithm. However, this kind of approach %meets
faces various challenges for our general problem where the convergence is both \emph{sublinear} and \emph{non-monotonic}. Since the convergence rate is sublinear (see Remark \ref{Remark sublinear}), showing a linear contraction per iteration is no longer possible. Since the convergence is non-monotonic\footnote{To see this, consider $n=2, \mu_1=0, \mu_2=(1,0,\ldots,0)^\top$, then the norm of $\mu_1$ strictly increases after one iteration.}, we also cannot show a strictly decreasing parametric distance.

To address these challenges, we propose a new convergence analysis framework for gradient EM by proving the convergence of \emph{likelihood} $\loss$ instead of the convergence of parameters $\bmu$. There are several benefits for considering the convergence from the perspective of MLE loss $\loss$. Firstly, it naturally addresses the problem of non-monotonic and sub-linear convergence since we only need to show $\loss$ decreases as the algorithm updates. Also, since gradient EM is equivalent with running gradient descent on loss function $\loss$ (see Section \ref{gradient EM = GD}), we can apply techniques from the optimization theory of gradient descent to facilitate our analysis.

\subsection{Proof ideas for Theorem \ref{Global convergence of gradient EM}}

We first briefly outline our proof of Theorem \ref{Global convergence of gradient EM}.

\bftext{Proof roadmap.}
Our proof of Theorem \ref{Global convergence of gradient EM} consists of three steps. Firstly, we prove a gradient lower bound for $\loss$ (Theorem \ref{Gradient projection lower bound}). Then we prove that the MLE $\loss$ is \emph{locally smooth} (Theorem \ref{smoothness}). Finally, we combine the gradient lower bound and the smoothness condition to prove the global convergence of $\loss$ with mathematical induction.

\bftext{Step 1: Gradient lower bound.}

Our first step aims to show that the gradient norm of $\loss(\bmu)$ is lower bounded by the distance of $\bmu$ to the ground truth. To do this, we need a few preliminary results. Inspired by \citet{chen2023local}, we use Stein's identity \citep{stein81} to perform an algebraic transformation of the gradient. Recalling the definition of $\psi_i$ in \eqref{psi i definition}, we have the following lemma.

\begin{lemma}\label{Gradient transformation lemma}
    For any $\GMM(\bmu), i\in[n]$, the gradient of $Q$ satisfies
    \[\nabla_{\mu_i}\loss(\bmu)= \nabla_{\mu_i} Q(\bmu|\bmu)=\E_x\left[\psi_i(x)\sum_{k\in[n]}\psi_k(x)\mu_k\right].\]
\end{lemma}

The gradient expression above is equivalent with the form in \eqref{gradient EM update}, but is easier to manipulate. Using the transformed gradient in Lemma \ref{Gradient transformation lemma}, we have the following corollary.

\begin{corollary}\label{Gradient projection lemma} 
Define vector $\tpsi_{\bmu}(x)\coloneqq \sum_{i\in[n]}\psi_i(x)\mu_i$.
For any $\GMM(\bmu)$, the projection of the gradient of $\nabla \loss(\bmu)$ onto $\bmu$ satisfies
\[\ip{\nabla \loss(\bmu), \bmu}=\ip{\nabla_{\bmu} Q(\bmu|\bmu), \bmu}=\sum_{i\in[n]}\ip{\nabla_{\mu_i}Q(\bmu|\bmu), \mu_i} = \E_x\left[\left\|\tpsi_{\bmu}(x)\right\|^2\right].\]
    
\end{corollary}

Corollary \ref{Gradient transformation lemma} is important since it converts the projection of gradient $\nabla\loss(\bmu) $ onto $\bmu$ to the expected norm square of a vector $\tpsi_{\bmu}$. Since a lower bound of the gradient projection implies a lower bound of the gradient, we only need to construct a lower bound for $\ip{\nabla \loss(\bmu), \bmu}= \E_x\left[\left\|\tpsi_{\bmu}(x)\right\|^2\right]$. Since $\left\|\tpsi_{\bmu}(x)\right\|^2$ is always non-negative, we already know that the gradient projection is non-negative. But lower bounding $\E_x\left[\left\|\tpsi_{\bmu}(x)\right\|^2\right]$ is still highly nontrivial since the expression of $\tpsi$ is complicated and hard to handle. However, our key observation is that, \emph{although $\tpsi$ itself is hard to bound, its gradient has nice properties and can be handled gracefully}:
\begin{equation}\label{main text gradient of tpsi}
        \nabla_x \tpsi_{\bmu}(x)
        =\frac{1}{2}\sum_{i,j\in[n]} \psi_i(x)\psi_j(x)(\mu_i-\mu_j)(\mu_i-\mu_j)^{\top}.
\end{equation}
The gradient \eqref{main text gradient of tpsi} is nicely-behaved. One can see immediately from \eqref{main text gradient of tpsi} that the matrix $\nabla_x \tpsi_{\bmu}(x)$ is positive-semi-definite, and its eigenvalues can be directly bounded. To utilize these properties, we use the following algebraic trick to convert the task of lower bounding $\tpsi$ itself into the task of lower bounding its gradient.

\begin{equation}\label{main text tpsi to gradient}
    \E_x\left[\|\tpsi_{\bmu}(x)\|^2\right]=\frac{1}{4}\E_x\left[\left(\int_{t=-1}^{1} \|x\|\cdot\overline{x}^{\top}\nabla \tpsi_{\bmu}(tx)\overline{x}\diff t\right)^2\right].
\end{equation}
Recall that $\bar{x}=\frac{x}{\|x\|}$. See detailed derivation in \eqref{tpsi to gradient}.
Using \eqref{main text gradient of tpsi}, combined with the properties of $\nabla_x \tpsi_{\bmu}(x)$, we can obtain the following lemma (Recall that $U=\sum_{i\in[n]} \|\mu_i\|^2$.):

\begin{lemma}\label{separation lower bound}
    For any $\GMM(\bmu)$ we have
    \[ \E_x\left[\|\tpsi_{\bmu}(x)\|^2\right]\geq \frac{\exp\left(-8U\right)}{40000 d(1+2\mumax{\sqrt{d}})^2}\left(\sum_{i,j\in[n]} \pi_i\pi_j\|\mu_i-\mu_j\|^2\right)^2.\]
\end{lemma}

On top of Lemma \ref{separation lower bound}, we can easily lower bound the gradient projection in the following lemma, finishing the first step of our proof.
\begin{lemma}[Gradient projection lower bound] \label{Gradient projection lower bound}
    For any $\GMM(\bmu)$ we have
\[\ip{\nabla_{\bmu} Q(\bmu|\bmu), \bmu}=\E_x[\|\tpsi_{\bmu}(x)\|^2] = \Omega\left(\frac{\exp\left(-8U\right)\pimin^2}{d(1+\mumax{\sqrt{d}})^2}\mumax^4\right).\]
\end{lemma}

\bftext{Step 2: Local smoothness.}

To construct a global convergence analysis for gradient-based methods, after obtaining a gradient lower bound, we still need to prove the smoothness of loss $\loss$. (Recall that global smoothness of function $f$ means that there exists constant $C$ such that $\|\nabla f(x_1)-\nabla f(x_2)\|\leq C\|x_1-x_2\|, \forall x_1, x_2$.) However, proving the smoothness for $\loss$ in general is very challenging since the membership function $\psi_i$ cannot be bounded when $\bmu$ is  unbounded. To address this issue, we prove that $\loss$ is locally smooth, $i.e.$, the smoothness between two points $\bmu$ and $\bmu'$ is satisfied if both $\|\bmu\|$ and $\|\bmu-\bmu'\|$ are upper bounded. Our result is the following theorem.
\begin{theorem}[Local smoothness of loss function]\label{smoothness}
At any two points $\bm{\mu} =(\mu_1^{\top}, \ldots, \mu_n^{\top})^{\top}$ and $\bmu + \bm{\delta}=((\mu_1+\delta_1)^{\top}, \ldots, (\mu_n+\delta_n)^{\top})^{\top}$, if 
\[\|\delta_i\|\leq \frac{1}{\max\left\{6d,2\|\mu_i\|\right\}}, \forall i\in[n],\] 
then the loss function $\loss$ satisfies the following smoothness property: for any $i\in[n]$ we have
\begin{equation}
   \left\|\nabla_{\mu_i+\delta_i}\loss (\bmu+\bm{\delta})-\nabla_{\mu_i}\loss (\bmu)\right\|\leq  n\mumax(30\sqrt{d}+4\mumax)\|\delta_i\|+\sum_{k\in[n]}\|\delta_k\|.
\end{equation}
\end{theorem}

\bftext{Step 3: putting everything together.}

Given the gradient lower bound and the smoothness condition, we still need to resolve two remaining problems. The first one is that the gradient lower bound in Lemma \ref{Gradient projection lower bound} is given in terms of $\bmu$, which we need to convert to a lower bound in terms of $\loss(\bmu)$. For this we need the following upper bound of $\loss$.

\begin{theorem}[Loss function upper bound]\label{Loss function upper bound}
    The loss function can be upper bounded as
     \[\loss(\bmu)\leq \sum_{i\in[n]}\frac{\pi_i}{2}\|\mu_i\|^2\leq \frac{\mumax^2}{2}.\]
\end{theorem}

The second problem is that our local smoothness theorem requires $\bmu$ to be bounded, therefore we need to show a regularity condition that for each $i$, $\bmu_i(t)$ stays in a bounded region during gradient EM updates. This is not easy to prove for each individual $\bmu_i$ due to the same non-monotonic issue mentioned in Section \ref{Section Hardness}. To establish such a regularity condition, we use the potential function.$U$ to solve this problem.
We prove that $U$ remains bounded along the gradient EM trajectory, implying %the regularity condition for 
each $\bmu_i$ remains well-behaved. With this regularity condition, combined with the previous two steps, we finish the proof of Theorem \ref{Global convergence of gradient EM} via mathematical induction.

% \begin{theorem}[Loss function lower bound]
%     The loss function can be lower bounded as
%     \[\loss(\bmu)\geq \]
% \end{theorem}

% \begin{proof}
% Since the logarithm function is concave, by Jensen's inequality we have
%     \[\begin{split}
%         \loss(\bmu) &= \kl(p_{\bmu^*}||p_{\bmu})=-\E_{x}\left[\log\left(\frac{p_{\bmu}(x)}{p_{\bmu^*}(x)}\right)\right]\\
%         &=-\E_{x}\left[\log\left(\frac{\sum_{i} \pi_i\exp\left(-\frac{\|x-\mu_i\|^2}{2}\right)}{\exp\left(-\frac{\|x\|^2}{2}\right)}\right)\right]
%     \end{split}\]
% \end{proof}

\subsection{Proof ideas for Theorem \ref{Existence of bad initialization region}}
Proving Theorem \ref{Existence of bad initialization region} is much simpler. The idea is natural: we found that there exists some bad regions where the gradient of $\loss$ is exponentially small, characterized by the following lemma.

\begin{lemma}[Gradient norm upper bound]\label{Gradient norm upper bound}
For any $\bmu$ satisfying $\|\mu_1\|,\|\mu_2\|\geq 10\sqrt{d}, \|\mu_3\|,\ldots,\|\mu_n\|\leq \sqrt{d}$, the gradient of $\loss$ at $\bmu$ can be upper bounded as
    \[\|\nabla_{\mu_i}\loss(\bmu)\| \leq 2(\|\mu_3\|+\cdots+\|\mu_n\|)+2\exp(-d)(\|\mu_1\|+\|\mu_2\|),\forall i\in[n].\]
\end{lemma}

Utilizing Lemma \ref{Gradient norm upper bound}, we can prove Theorem \ref{Existence of bad initialization region} by showing that initialization from these bad regions will get trapped in it for exponentially long, since the gradient norm is exponentially small. The full proof can be found in Appendix \ref{Proofs for counter-example theorem}.

\section{Experiments}
\begin{figure}[t]
    \begin{minipage}{0.325\linewidth}
    \centering
    \includegraphics[width=\linewidth]{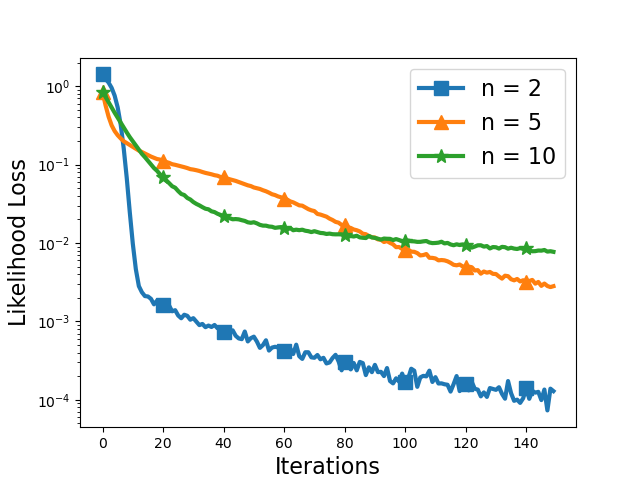 }
    % \captionof{figure}{\label{fig likelihood}}
     \end{minipage}
     \hfill
     \begin{minipage}{0.325\linewidth}
    \centering
    \includegraphics[width=\linewidth]{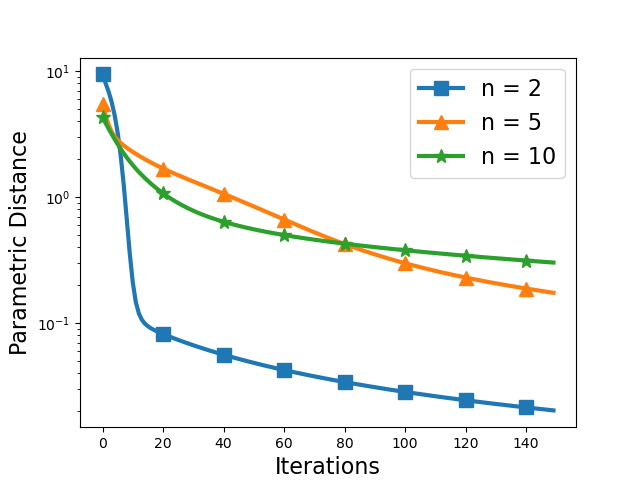 }
    % \captionof{figure}{\label{fig parametric} }
     \end{minipage}
     \hfill
      \begin{minipage}{0.325\linewidth}
    \centering
    \includegraphics[width=\linewidth]{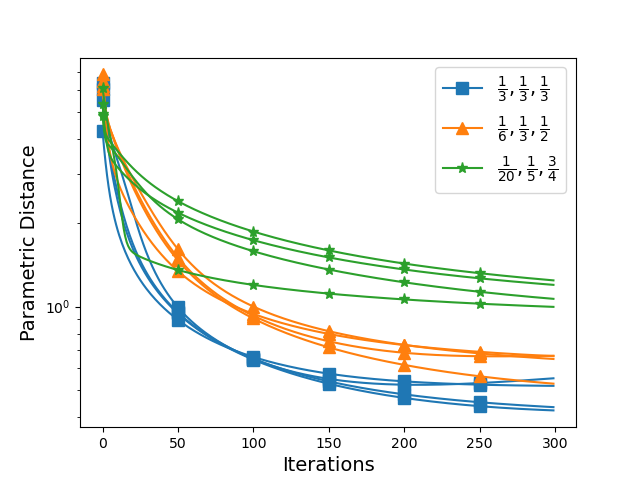 }
    % \captionof{figure}{\label{fig counter-example}}
     \end{minipage}
     
     \caption{\label{simulation}
 Left: Sublinear convergence of the likelihood loss $\loss$. Middle: Sublinear convergence of the parametric distance $\sum_{i\in[n]}\pi_i\|\mu_i-\mu^*\|^2 $ between student GMM and the ground truth. Right: Impact of different mixing weights on the convergence speed.}
\end{figure}

\begin{figure}[t]
\centering
    \begin{minipage}{0.4\linewidth}
    \centering
    \includegraphics[width=\linewidth]{ 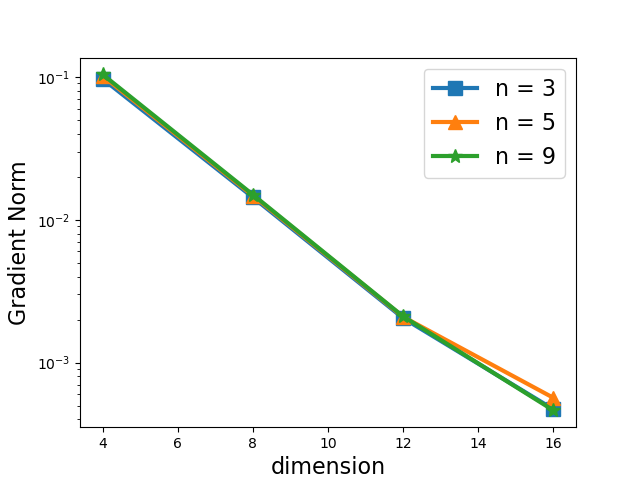 }
    % \captionof{figure}{\label{fig likelihood}}
     \end{minipage}
      \begin{minipage}{0.4\linewidth}
    \includegraphics[width=\linewidth]{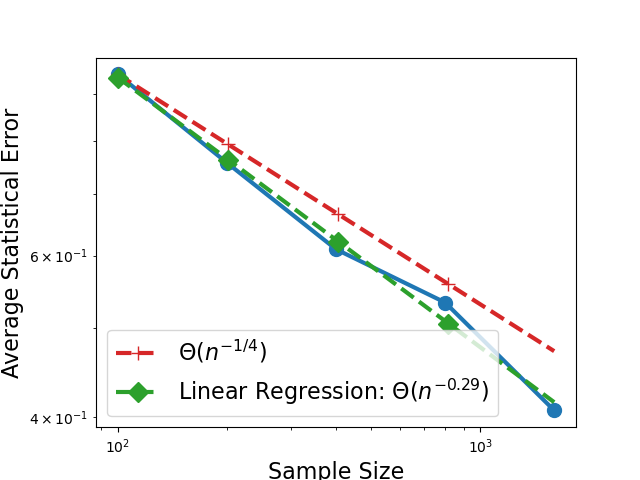}
    % \captionof{figure}{\label{fig counter-example}}
     \end{minipage}
     
     \caption{\label{simulation-2}
 Left: Gradient norm $\|\nabla \loss(\bmu(0))\|$ in the counter-example in Theorem \ref{Existence of bad initialization region} decreases exponentially fast w.r.t. dimension $d$. Right: The statistical error (blue line) approximately scales as $\sim n^{-1/4}$ with sample size $n$.
     }
\end{figure}

In this section we experimentally explore the behavior of gradient EM on GMMs.

\bftext{Convergence rates.} We choose the experimental setting of $d=5, \eta = 0.7$. We use $n=2,5,10$ Gaussian mixtures to learn data generated from one single ground truth Gaussian distribution $\N(\mu^*,I_d)$, respectively. Since a closed form expression of the population gradient is intractable, we approximate the gradient step via Monte Carlo method, with sample size $3.5\times 10^5$.  The mixing weights of student GMM are randomly sampled from a standard Dirichlet distribution and set as fixed during gradient EM update. The covariances of all component Gaussians are set as the identity matrix.  We recorded the convergence of likelihood function $\loss$ (estimated also by  Monte Carlo method on fresh samples each iteration) and parametric distance $\sum_{i\in[n]}\pi_i\|\mu_i-\mu^*\|^2 $ along gradient EM trajectory. The results are reported in Figure \ref{simulation} (left and middle panel). Both the likelihood $\loss$ and the parametric distance converges sub-linearly.

\bftext{Weight configurations.} We train $3$-component GMM with $3$-different weight configurations and report $4$ runs each configuration in Figure \ref{simulation} (right). Blue: $(\frac{1}{3}, \frac{1}{3}, \frac{1}{3})$. Orange: $(\frac{1}{6}, \frac{1}{3}, \frac{1}{2})$, Green: $(\frac{1}{20}, \frac{1}{5}, \frac{3}{4})$. More evenly distributed weights result in faster convergence.

\bftext{Initialization geometry.} We empirically study the bad initialization point $\bmu(0)$ described in Theorem \ref{Existence of bad initialization region} \footnote{To prevent numerical underflow issues, we change the constant $12$ in $\bmu(0)$ to $2$.} by plotting the gradient norm at $\bmu(0)$ w.r.t. different dimension $d$ in Figure \ref{simulation-2} (left). As theoretically analyzed, the gradient norm $\|\nabla \loss(\bmu(0))\|$ at $\bmu(0)$ decreases exponentially in dimension $d$.

\bftext{Statistical rates.} The statistical rate for EM/gradient EM is another interesting research problem, which we observe empirically in Figure \ref{simulation-2} (right).
We run gradient EM on $5$-component GMM with equal weights. x-axis: number of training samples, y-axis: parametric error after convergence.
     For each sample size, we run  $50$ times and report the average. The statistical errors are reported in the blue line. The red line (function $\Theta(n^{-1/4})$) and green line (linear regression output fitting blue points) are references. The trajectory approximately follows the law of $accuracy \propto n^{-1/4}$. While \citep{wu2019randomly} rigorously proves the asymptotic statistical rate of $\Tilde{O}(n^{-1/4})$ for the special case of 2-GMMs, our experiments imply that the same rate might also apply to the general case of multi-component GMMs.

% \begin{lemma}
%     For any $t>0$ we have
%     \[\frac{1-e^{-t}}{t}\geq \frac{1}{1+t}.\]
% \end{lemma}

% \begin{proof}
% Since $t>0$, we have $e^t\geq 1+t \To e^{-t}\leq \frac{1}{1+t}$, therefore
%     \[\frac{1-e^{-t}}{t}\geq \frac{1-\frac{1}{1+t}}{t}=\frac{1}{1+t}.\]
% \end{proof}

% \begin{lemma}
%     For any vector $v\in\R^n$, the following holds:
%     \[\]
%     \[\int_{\|x\|\leq 2\sqrt{d}} \|x\|\langle v, \overline{x}\rangle^2 \phi(x|0,I)\diff x\geq \Omega(\|v\|^2/\sqrt{d}).\]
% \end{lemma}

% \begin{proof}
%     \begin{align*}
%         \int_{\|x\|\leq 2\sqrt{d}} \|x\|\langle v, \overline{x}\rangle^2 \phi(x|0,I)\diff x&=\frac{\|v\|^2}{d}\int_{r=0}^{2\sqrt{d}}r(2\pi)^{-d/2}\exp(-r^2/2)\frac{2\pi^{d/2}r^{d-1}}{\Gamma(\frac{d}{2})}\diff r\\
%         &=\frac{\|v\|^2}{d}\left(\sqrt{2}\frac{\Gamma(\frac{d+1}{2})}{\Gamma(\frac{d}{2})}-\int_{2\sqrt{d}}^{+\infty}r(2\pi)^{-d/2}\exp(-r^2/2)\frac{2\pi^{d/2}r^{d-1}}{\Gamma(\frac{d}{2})}\diff r\right)\\
%         &=\frac{\|v\|^2}{d}\sqrt{2}\frac{\gamma(\frac{d+1}{2}, 2d)}{\Gamma(\frac{d}{2})}\\.
%     \end{align*}
% (We have $\Gamma((d+1)/2)\geq \sqrt{d/2}\Gamma(d/2)$ due to the convexity of $\ln\Gamma$.)

% In fact, this is equivalent with a tail bound of the Chi distribution.

% \end{proof}

%%%%%%%%%%%%%%%%%%%%%%%%%%%%%%%%%%%%%%%%%%%%%%%%%%%%%%%%%%%%
\section{Conclusion}
This paper gives the first global convergence of gradient EM for over-parameterized Gaussian mixture models when the ground truth is a single Gaussian, and rate is sublinear which is exponentially slower than the rate in the exact-parameterization case.
One fundamental open problem is to study when one can obtain global convergence of EM or gradient EM for Gaussian mixture models when the ground truth has multiple components. The likelihood-based convergence framework proposed in this paper might be an helpful tool towards solving this general problem.

\paragraph{Acknowledgements} This work was supported in part  by the following grants: NSF TRIPODS II-DMS 20231660, NSF CCF 2212261, NSF CCF 2007036, NSF AF 2312775, NSF IIS 2110170, NSF DMS 2134106, NSF IIS 2143493, and NSF IIS 2229881.

\bibliography{ref}

% \section{Appendix / supplemental material}
% Optionally include supplemental material (complete proofs, additional experiments and plots) in appendix.
% All such materials \textbf{SHOULD be included in the main submission.}

%%%%%%%%%%%%%%%%%%%%%%%%%%%%%%%%%%%%%%%%%%%%%%%%%%%%%%%%%%%%
\appendix
\newpage
\section{Missing Proofs and Auxiliary lemmas}

\begin{proof}[Proof of Fact~\ref{Fact}]
    It is well known that (see Section 1 of \cite{wu2019randomly})
    \[Q(\bmu'|\bmu)= \E_{x\sim p_{\bmu^*}}\left[\log(p_{\bmu'}(x))-\kl(p_{\bmu}(\cdot|x)||p_{\bmu'}(\cdot|x))-H(p_{\bmu}(\cdot|x))\right],\]
    where $ p_{\bmu}(\cdot|x)$ denotes the distribution of hidden variable $y$ (in our case of GMM the index of Gaussian component) conditioned on $x$, and $H$ denotes information entropy.

    Since $\bmu'=\bmu$ is a global minimum of $\kl(p_{\bmu}(\cdot|x)||p_{\bmu'}(\cdot|x))$, we have $\nabla\kl(p_{\bmu}(\cdot|x)||p_{\bmu}(\cdot|x))=0$. Also $\nabla H(p_{\bmu}(\cdot|x))=0$ since $H(p_{\bmu}(\cdot|x))$ is a constant. Therefore
    \[\nabla Q(\bmu|\bmu)=\E_{x\sim p_{\bmu^*}}\left[\nabla\log(p_{\bmu}(x))\right]=\nabla\loss(\bmu).\]
\end{proof}

The proof of Lemma \ref{Gradient transformation lemma} uses ideas from Theorem 1 of \citet{chen2023local} and relies on Stein's identity, which is given by the following lemma.
\begin{lemma}[\citet{stein81}]\label{Stein's identity}
For $x\sim \N(\mu,\sigma^2 I_d)$ and differentiable function $g:\R^d\to \R$ we have
\[\E[g(x)(x-\mu)]=\sigma^2\E[\nabla_x g(x)],\]
if the two expectations in the above identity exist.
\end{lemma}

% \begin{proof}
%     \begin{align*}
%         \E[g(x)(x-\mu)]&=(2\pi\sigma^2)^{-d/2}\int_{\R^d} g(x)(x-\mu)\exp\left(-\frac{\|x-\mu\|^2}{2\sigma^2}\right)\diff x\\
%         &=(2\pi\sigma^2)^{-d/2}\int_{\R^d} -\sigma^2 g(x)\diff \exp\left(-\frac{\|x-\mu\|^2}{2\sigma^2}\right)\\
%         &=(2\pi\sigma^2)^{-d/2}\int_{\R^d} \sigma^2  \exp\left(-\frac{\|x-\mu\|^2}{2\sigma^2}\right)\diff g(x)\\
%         &=\sigma^2\E[\nabla g(x)].
%     \end{align*}
% \end{proof}

Now we are ready to prove Lemma \ref{Gradient transformation lemma}.
\begin{replemma}{Gradient transformation lemma}
For any $\GMM(\bmu), i\in[n]$, the gradient of $Q$ satisfies
    \[\nabla_{\mu_i}\loss(\bmu)=\nabla_{\mu_i} Q(\bmu|\bmu)=\E_x\left[\psi_i(x)\sum_{k\in[n]}\psi_k(x)\mu_k\right].\]
\end{replemma}

\begin{proof}
Applying Stein's identity (Lemma \ref{Stein's identity}), for each $i\in[n]$ we have
    \[\begin{split}
        \nabla_{\mu_i} Q(\bm{\mu}|\bm{\mu})&= \E_{x\sim \N(0,I_d)} \left[\-\psi_i(x)(\mu_i-x)\right]\\
        &=\E_{x\sim \N(0,I_d)} \left[\-\psi_i(x)\right]\mu_i-\E_{x\sim \N(0,I_d)} \left[\psi_i(x)x\right]\\
        &=\E_{x\sim \N(0,I_d)} \left[\-\psi_i(x)\right]\mu_i-\E_{x\sim \N(0,I_d)}[\nabla_x\psi_i(x)].
    \end{split}\]
Recall that 
\[\psi_i(x) =\Pr [i|x]=\frac{\pi_i\exp\left(-\frac{\|x-\mu_i\|^2}{2}\right)}{\sum_{k\in[n]} \pi_k\exp\left(-\frac{\|x-\mu_k\|^2}{2}\right)}.\]
The gradient $\nabla_x\psi_i(x)$ could be calculated as
\begin{equation}\label{psi gradient}
 \begin{split}
     &\nabla_x \psi_i(x)\\
     ={}& \frac{1}
     {\left(\sum_{k\in[n]} \pi_k\exp\left(-\frac{\|x-\mu_k\|^2}{2}\right)\right)^2}
     \Bigg[\left(\sum_{k\in[n]} \pi_k\exp\left(-\frac{\|x-\mu_k\|^2}{2}\right)\right)\pi_i\exp\left(-\frac{\|x-\mu_i\|^2}{2}\right)(\mu_i-x)\\
     &-\pi_i\exp\left(-\frac{\|x-\mu_i\|^2}{2}\right)\left(\sum_{k\in[n]} \pi_k\exp\left(-\frac{\|x-\mu_k\|^2}{2}\right)(\mu_k-x)\right)\Bigg]\\
 ={}&\psi_i(x)(\mu_i-x)-\psi_i(x)\sum_{k\in[n]}\psi_k(x)(\mu_k-x)\\
 ={}&\psi_i(x)(\mu_i-x)+\psi_i(x)x-\sum_{k\in[n]}\psi_i(x)\psi_k(x)\mu_k\\
 ={}&\psi_i(x)\left(\mu_i-\sum_{k\in[n]}\psi_k(x)\mu_k\right),
 \end{split}
\end{equation}
note that we used $\sum_{k\in[n]} \psi_i(x) =1$.

Then we have
\[\begin{split}
     \nabla_{\mu_i} Q(\bm{\mu}|\bm{\mu})&=\E_{x} \left[\-\psi_i(x)\right]\mu_i-\E_{x}[\nabla_x\psi_i(x)]\\
     &=\E_{x} \left[\-\psi_i(x)\right]\mu_i-\E_{x}\left[\psi_i(x)\left(\mu_i-\sum_{k\in[n]}\psi_k(x)\mu_k\right)\right]=\E_x\left[\psi_i(x)\sum_{k\in[n]}\psi_k(x)\mu_k\right].
\end{split}\]
\end{proof}

% \simon{move the proof to appendix.}
\begin{proof}[Proof of Corollary~\ref{Gradient projection lemma}]
    \[\begin{split}
    &\ip{\nabla_{\bmu} Q(\bmu|\bmu), \bmu}=\sum_{i\in[n]}\ip{\nabla_{\mu_i}Q(\bmu|\bmu), \mu_i}
    =\sum_{i\in[n]}\ip{\E_x\left[\psi_i(x)\sum_{k\in[n]}\psi_k(x)\mu_k\right],\mu_i}\\
    &=\sum_{i\in[n]}\sum_{k\in[n]}\E_x\ip{\psi_i(x)\psi_k(x)\mu_k,\mu_i}
    =\E_x\left[\left\|\sum_{i\in[n]}\psi_i(x)\mu_i\right\|^2\right]= \E_x\left[\left\|\tpsi_{\bmu}(x)\right\|^2\right].
\end{split}\]

\end{proof}

\begin{lemma}\label{MGF upper bound}
    For any constant $c$ satisfying $0<c\leq \frac{1}{3d}$, we have
    \[\E_{x\sim\N(0,I_d)} \left[\exp\left(c\|x\|\right)\right]\leq 1+5\sqrt{d}c.\]
\end{lemma}

\begin{proof}

Note that $\E_{x\sim\N(0,I_d)} \left[\exp\left(c\|x\|\right)\right]=\mgf(c)$ is the moment-generating function of $\|x\|$. To upper bound the value of a moment generating function at $c$, we use Lagrange's Mean Value Theorem:
\begin{equation}\label{NL13}
    \mgf(c)=\mgf(0)+\mgf'(\xi)c,
\end{equation}
where $\xi\in[0,c]$. Note that 
$\mgf(0)=1,$
So the remaining task is to bound $\mgf'(\xi)$.
We bound this expectation using truncation method as:
\begin{equation}\label{NL12}
    \begin{split}
        \mgf'(\xi)&=\E_x\left[\|x\|\exp(\xi\|x\|)\right]\leq\E_x\left[\|x\|\exp(c\|x\|)\right] \\
        &=\int_{x\in\R^d}\|x\|\exp(c\|x\|)(2\pi)^{-d/2}\exp\left(-\frac{\|x\|^2}{2}\right)\diff x\\
        &=\int_{\|x\|\leq 1}\|x\|\exp(c\|x\|)(2\pi)^{-d/2}\exp\left(-\frac{\|x\|^2}{2}\right)\diff x\\
        &\quad +\int_{\|x\|\geq 1}\|x\|\exp(c\|x\|)(2\pi)^{-d/2}\exp\left(-\frac{\|x\|^2}{2}\right)\diff x\\
        &\leq \exp(c)(2\pi)^{-d/2}V_d+\int_{\|x\|\geq 1}\|x\|(2\pi)^{-d/2}\exp\left(c\|x\|-\frac{\|x\|^2}{2}\right)\diff x\\
        &\leq \exp(c)(2\pi)^{-d/2}V_d+\int_{\|x\|\geq 1}\|x\|(2\pi)^{-d/2}\exp\left(c\|x\|-\frac{\|x\|^2}{2}\right)\diff x,
    \end{split}
\end{equation}
where $V_d = \frac{\pi^{d/2}}{\Gamma(d/2+1)}$ is the volume of $d$-dimensional unit sphere.

Since $\|x\|\geq 1\To c\|x\|-\frac{\|x\|^2}{2}\leq \frac{1}{3d}\|x\|-\frac{\|x\|^2}{2}\leq -\frac{\|(1-1/(2d))x\|^2}{2}$, we have
\[\begin{split}
    &\quad\int_{\|x\|\geq 1}\|x\|(2\pi)^{-d/2}\exp\left(c\|x\|-\frac{\|x\|^2}{2}\right)\diff x\\
    &\leq \int_{\|x\|\geq 1}\|x\|(2\pi)^{-d/2}\exp\left(-\frac{\|\frac{2d-1}{2d}x\|^2}{2}\right)\diff x\\
    &=\int_{\|y\|\geq \frac{2d-1}{2d}}\frac{2d}{2d-1}\|y\|(2\pi)^{-d/2}\exp\left(-\frac{\|y\|^2}{2}\right)\left(\frac{2d}{2d-1}\right)^d\diff y\\
    &\leq \left(\frac{2d}{2d-1}\right)^{d+1}\E_{y\sim\N(0,I_d)} \left[\|y\|\right]\\
    &=\left(\frac{2d}{2d-1}\right)^{d+1}\frac{\sqrt{2}\Gamma\left(\frac{d+1}{2}\right)}{\Gamma\left(\frac{d}{2}\right)}\\
    &\leq 4\sqrt{d},
\end{split}\]
where we used $\left(\frac{2d}{2d-1}\right)^{d+1}\leq 4$ and the log convexity of Gamma function at the last line. Plugging this back to \eqref{NL12}, we get

\begin{equation}\label{NL14}
    \begin{split}
        \mgf'(\xi)
        &\leq \exp(c)(2\pi)^{-d/2}V_d+\int_{\|x\|\geq 1}\|x\|(2\pi)^{-d/2}\exp\left(c\|x\|-\frac{\|x\|^2}{2}\right)\diff x\\
        &\leq \exp(1/(3d))(2\pi)^{-d/2}+4\sqrt{d}\\
        &\leq 5\sqrt{d}.
    \end{split}
\end{equation}
Plugging \eqref{NL14} into \eqref{NL13}, we obtain the final bound
\[\E_x\left[\exp\left(2\|\delta_i\|(\|x\|+\|\mu_i\|)\right)-1\right]=\mgf(c)=\mgf(0)+\mgf'(\xi)c\leq 1+5\sqrt{d}c.\]
\end{proof}

\begin{lemma}\label{integral psi(tx) bound lemma}
    Recall that $U=\sum_{i\in[n]}\|\mu_i\|^2$. For any fixed $x\in \R^d, x\neq 0$ and any $\bmu$ we have
    \[\int_{t=-1}^{1}\psi_i(tx|\bmu)\psi_j(tx|\bmu) \diff t\geq\frac{1}{2\mumax\|x\|}\pi_i\pi_j\exp\left(-4U\right)\left(1-\exp\left(-4\mumax\|x\|\right)\right).\]
\end{lemma}

\begin{proof}
    \begin{equation}
        \begin{split}
            \psi_i(tx) &= \frac{\pi_i\exp\left(-\frac{\|tx-\mu_i\|^2}{2}\right)}{\sum_{k\in[n]} \pi_k\exp\left(-\frac{\|tx-\mu_k\|^2}{2}\right)}\\
            &= \frac{\pi_i}{\sum_{k\in[n]} \pi_k\exp\left(\frac{1}{2}(\|tx-\mu_i\|^2-\|tx-\mu_k\|^2)\right)}\\
            &= \frac{\pi_i}{\sum_{k\in[n]} \pi_k\exp\left(\frac{1}{2}(\|tx-\mu_i\|^2-\|tx-\mu_k\|^2)\right)}\\
            &= \frac{\pi_i}{\sum_{k\in[n]} \pi_k\exp\left(\frac{1}{2}\ip{2tx-\mu_i-\mu_k, \mu_k-\mu_i}\right)}\\
            &\geq \frac{\pi_i}{\sum_{k\in[n]} \pi_k\exp\left(\frac{1}{2}(2\|tx\|+2\mu_{\max})\cdot 2\mumax \right)}\\
            &=\pi_i\exp\left(-2\mumax(\|tx\|+\mumax)\right)
        \end{split}
    \end{equation}

Therefore
    \begin{equation}
        \begin{split}
            \int_{t=-1}^{1}\psi_i(tx)\psi_j(tx) \diff t &\geq \int_{t=-1}^{1}\pi_i\pi_j \exp\left(-4\mumax(\|tx\|+\mumax)\right) \diff t\\
            &= \pi_i\pi_j\exp\left(-4\mumax^2\right)\cdot 2\int_{t=0}^{1} \exp\left(-4\mumax\|x\|t\right) \diff t\\
            &\geq\frac{1}{2\mumax\|x\|}\pi_i\pi_j\exp\left(-4U\right)\left(1-\exp\left(-4\mumax\|x\|\right)\right).
        \end{split}
    \end{equation}

\end{proof}

\section{Proofs for Section \ref{Section Main Result} and \ref{Section proof overview}}

\subsection{Proofs for global convergence analysis}

\begin{reptheorem}{smoothness}
At any two points $\bm{\mu} =(\mu_1^{\top}, \ldots, \mu_n^{\top})^{\top}$ and $\bmu + \bm{\delta}=((\mu_1+\delta_1)^{\top}, \ldots, (\mu_n+\delta_n)^{\top})^{\top}$, if 
\[\|\delta_i\|\leq \frac{1}{\max\left\{6d,2\|\mu_i\|\right\}}, \forall i\in[n],\] 
then the loss function $\loss$ satisfies the following smoothness property: for any $i\in[n]$ we have
\begin{equation}
   \left\|\nabla_{\mu_i+\delta_i}\loss (\bmu+\bm{\delta})-\nabla_{\mu_i}\loss (\bmu)\right\|\leq  n\mumax(30\sqrt{d}+4\mumax)\|\delta_i\|+\sum_{k\in[n]}\|\delta_k\|.
\end{equation}
    
\end{reptheorem}

\begin{proof}

Note that
\[\begin{split}
    &\quad\exp\left(-\|\delta_i\|(\|x\|+\|\mu_i\|)\right)\exp\left(-\frac{\|\delta_i\|^2}{2}\right)\leq \frac{\exp\left(-\frac{\|x-(\mu_i+\delta_i)\|^2}{2}\right)}{\exp\left(-\frac{\|x-\mu_i\|^2}{2}\right)}=\exp\left(\ip{x-\mu_i, \delta_i}-\frac{\|\delta_i\|^2}{2}\right)\\
    &\leq \exp\left(\|\delta_i\|(\|x\|+\|\mu_i\|)\right)\exp\left(-\frac{\|\delta_i\|^2}{2}\right).
\end{split}\]

Therefore $\psi_i(x|\bmu+\bm{\delta})$ can be bounded as
\begin{equation}\label{NL10}
    \begin{split}
    &\quad\psi_i(x|\bmu+\bm{\delta})=\frac{\pi_i\exp\left(-\frac{\|x-(\mu_i+\delta_i)\|^2}{2}\right)}{\sum_{k\in[n]} \pi_k\exp\left(-\frac{\|x-(\mu_k+\delta_k)\|^2}{2}\right)}\\
&\leq \frac{\pi_i\exp\left(-\frac{\|x-\mu_i\|^2}{2}\right)\exp\left(\|\delta_i\|(\|x\|+\|\mu_i\|)\right)\exp\left(-\frac{\|\delta_i\|^2}{2}\right)}{\sum_{k\in[n]} \pi_k\exp\left(-\frac{\|x-\mu_k\|^2}{2}\right)\exp\left(-\|\delta_i\|(\|x\|+\|\mu_i\|)\right)\exp\left(-\frac{\|\delta_i\|^2}{2}\right)}\leq \exp\left(2\|\delta_i\|(\|x\|+\|\mu_i\|)\right)\psi_i(x|\bmu).
\end{split}
\end{equation}

Similarly, we have
\begin{equation}\label{NL11}
    \begin{split}
    &\quad\psi_i(x|\bmu+\bm{\delta})=\frac{\pi_i\exp\left(-\frac{\|x-(\mu_i+\delta_i)\|^2}{2}\right)}{\sum_{k\in[n]} \pi_k\exp\left(-\frac{\|x-(\mu_k+\delta_k)\|^2}{2}\right)}\\
&\geq \frac{\pi_i\exp\left(-\frac{\|x-\mu_i\|^2}{2}\right)\exp\left(-\|\delta_i\|(\|x\|+\|\mu_i\|)\right)\exp\left(-\frac{\|\delta_i\|^2}{2}\right)}{\sum_{k\in[n]} \pi_k\exp\left(-\frac{\|x-\mu_k\|^2}{2}\right)\exp\left(\|\delta_i\|(\|x\|+\|\mu_i\|)\right)\exp\left(-\frac{\|\delta_i\|^2}{2}\right)}\geq \exp\left(-2\|\delta_i\|(\|x\|+\|\mu_i\|)\right)\psi_i(x|\bmu).
\end{split}
\end{equation}

Recall that by Lemma \ref{Gradient transformation lemma} we have
$\nabla_{\mu_i}\loss (\bmu) = \E_x\left[\psi_i(x|\bmu)\sum_{k\in[n]}\psi_k(x|\bmu)\mu_k\right],$ so
\begin{equation}\label{NL15}
    \begin{split}
        &\quad\left\|\nabla_{\mu_i+\delta_i}\loss (\bmu+\bm{\delta})-\nabla_{\mu_i}\loss (\bmu)\right\|\\ &= \left\|\E_x\left[\psi_i(x|\bmu+\bm{\delta})\sum_{k\in[n]}\psi_k(x|\bmu+\bm{\delta})(\mu_k+\delta_k)\right]-\E_x\left[\psi_i(x|\bmu)\sum_{k\in[n]}\psi_k(x|\bmu)\mu_k\right]\right\|\\
        &=\Bigg\|\E_x\left[\sum_{k\in[n]}\psi_i(x|\bmu+\bm{\delta})\psi_k(x|\bmu+\bm{\delta})\delta_k\right]\\
        &\quad+\E_x\left[\sum_{k\in[n]}(\psi_i(x|\bmu+\bm{\delta})\psi_k(x|\bmu+\bm{\delta})-\psi_i(x|\bmu)\psi_k(x|\bmu))\mu_k\right]\Bigg\|\\
        &\leq\E_x\left[\sum_{k\in[n]}\psi_i(x|\bmu+\bm{\delta})\psi_k(x|\bmu+\bm{\delta})\|\delta_k\|\right]\\
        &\quad+\E_x\left[\sum_{k\in[n]}|\psi_i(x|\bmu+\bm{\delta})\psi_k(x|\bmu+\bm{\delta})-\psi_i(x|\bmu)\psi_k(x|\bmu)|\cdot\|\mu_k\|\right]\\
        &\leq \sum_{k\in[n]}\|\delta_k\| +\sum_{k\in[n]}\E_x\left[|\psi_i(x|\bmu+\bm{\delta})\psi_k(x|\bmu+\bm{\delta})-\psi_i(x|\bmu)\psi_k(x|\bmu)|\right]\|\mu_k\|\\
        &\leq \sum_{k\in[n]}\|\delta_k\| +\sum_{k\in[n]}\E_x\left[\exp\left(2\|\delta_i\|(\|x\|+\|\mu_i\|)\right)-1\right]\|\mu_k\|,
     \end{split}
\end{equation}

where the last inequality is because $\psi_i,\psi_k\leq 1$ and applying \eqref{NL10} and \eqref{NL11}.

The remaining task is to bound $\E_x\left[\exp\left(2\|\delta_i\|(\|x\|+\|\mu_i\|)\right)-1\right]$. Since $2\|\delta_i\|\leq \frac{1}{3d}$, we can use Lemma \ref{MGF upper bound} to bound it as

\begin{equation}
    \begin{split}
        &\quad\E_x\left[\exp\left(2\|\delta_i\|(\|x\|+\|\mu_i\|)\right)-1\right]=\exp(2\|\delta_i\|\|\mu_i\|)\E_x\left[\exp\left(2\|\delta_i\|\cdot\|x\|)\right)\right]-1\\
            &\leq \exp(2\|\delta_i\|\|\mu_i\|)(1+10\sqrt{d}\|\delta_i\|)-1=\exp(2\|\delta_i\|\|\mu_i\|)-1+10\sqrt{d}\|\delta_i\|\exp(2\|\delta_i\|\|\mu_i\|)\\
        &\leq 4\|\delta_i\|\|\mu_i\|+10\sqrt{d}\|\delta_i\|\exp(1)\leq(30\sqrt{d}+4\|\mu_i\|)\|\delta_i\|.
    \end{split}
\end{equation}
where we used $\exp(1+x)\leq 1+2x, \forall x\in[0,1]$ at the last line. Plugging this back to \eqref{NL15}, we get
\begin{equation}
    \begin{split}
        &\quad\left\|\nabla_{\mu_i+\delta_i}\loss (\bmu+\bm{\delta})-\nabla_{\mu_i}\loss (\bmu)\right\|\\
        &\leq \sum_{k\in[n]}\|\delta_k\| +\sum_{k\in[n]}\E_x\left[\exp\left(2\|\delta_i\|(\|x\|+\|\mu_i\|)\right)-1\right]\|\mu_k\|\\
        &\leq \sum_{k\in[n]}\|\delta_k\| +\sum_{k\in[n]}(30\sqrt{d}+4\|\mu_i\|)\|\delta_i\|\|\mu_k\|\\
        &\leq n\mumax(30\sqrt{d}+4\mumax)\|\delta_i\|+\sum_{k\in[n]}\|\delta_k\| .
     \end{split}
\end{equation}
\end{proof}

\begin{reptheorem}{Loss function upper bound}
    The loss function can be upper bounded as
     \[\loss(\bmu)\leq \sum_{i\in[n]}\frac{\pi_i}{2}\|\mu_i\|^2\leq \frac{\mumax^2}{2}.\]
\end{reptheorem}

\begin{proof}
Since the logarithm function is concave, by Jensen's inequality we have
    \[\begin{split}
        \loss(\bmu) &= \kl(p_{\bmu^*}||p_{\bmu})=-\E_{x}\left[\log\left(\frac{p_{\bmu}(x)}{p_{\bmu^*}(x)}\right)\right]\\
        &=-\E_{x}\left[\log\left(\frac{\sum_{i} \pi_i\exp\left(-\frac{\|x-\mu_i\|^2}{2}\right)}{\exp\left(-\frac{\|x\|^2}{2}\right)}\right)\right]\\
        &\leq -\E_{x}\left[\sum_{i} \pi_i\log\left(\frac{\exp\left(-\frac{\|x-\mu_i\|^2}{2}\right)}{\exp\left(-\frac{\|x\|^2}{2}\right)}\right)\right]\\
        &=-\sum_{i} \pi_i\E_{x}\left[\ip{x, \mu_i}-\frac{\|\mu_i\|^2}{2}\right]\\
        &=\sum_{i\in[n]}\frac{\pi_i}{2}\|\mu_i\|^2\leq \frac{\mumax^2}{2}.
    \end{split}\]
\end{proof}

% \begin{theorem}[Loss function lower bound]
%     The loss function can be lower bounded as
%     \[\loss(\bmu)\geq \]
% \end{theorem}

% \begin{proof}
% Since the logarithm function is concave, by Jensen's inequality we have
%     \[\begin{split}
%         \loss(\bmu) &= \kl(p_{\bmu^*}||p_{\bmu})=-\E_{x}\left[\log\left(\frac{p_{\bmu}(x)}{p_{\bmu^*}(x)}\right)\right]\\
%         &=-\E_{x}\left[\log\left(\frac{\sum_{i} \pi_i\exp\left(-\frac{\|x-\mu_i\|^2}{2}\right)}{\exp\left(-\frac{\|x\|^2}{2}\right)}\right)\right]
%     \end{split}\]
% \end{proof}

\begin{replemma}{Gradient projection lower bound}
    For any $\GMM(\bmu)$ we have
\[\ip{\nabla_{\bmu} Q(\bmu|\bmu), \bmu}=\E_x[\|\tpsi_{\bmu}(x)\|^2] \geq \Omega\left(\frac{\exp\left(-8U\right)\pimin^2}{d(1+\mumax{\sqrt{d}})^2}\mumax^4\right).\]
\end{replemma}

\begin{proof}
    Consider two cases:
    
    \bftext{Case 1.} There exists $k\in[n]$ such that $\|\mu_k-\mu_{\imax}\|\geq \frac{\mumax}{2}$. Then by Lemma \ref{seperated mu} and Lemma \ref{separation lower bound} we have
    \[\begin{split}
        \E_x\left[\|\tpsi_{\bmu}(x)\|^2\right]&\geq \frac{\exp\left(-8U\right)}{40000 d(1+2\mumax{\sqrt{d}})^2}\left(\sum_{i,j\in[n]} \pi_i\pi_j\|\mu_i-\mu_j\|^2\right)^2\\
        &\geq \frac{\exp\left(-8U\right)}{40000 d(1+2\mumax{\sqrt{d}})^2}\left(\frac{\pimin}{8}\mumax^2\right)^2\\
        &=\frac{\exp\left(-8U\right)\pimin^2}{2560000 d(1+2\mumax{\sqrt{d}})^2}\mumax^4.
      \end{split}\]

    \bftext{Case2.} For $\forall k\in[n]$, $\|\mu_{\imax}-\mu_k\|< \frac{\mumax}{2}$. Then by Lemma \ref{almost parallel mu} we have $\E_x\left[\|\tpsi_{\bmu}(x)\|^2\right]\geq\frac{1}{4}\mumax^2\geq \Omega(\exp(-8\mumax^2)\mumax^4)\geq  \Omega(\exp(-8U)\mumax^4)\geq \Omega\left(\frac{\exp\left(-8U\right)\pimin^2}{d(1+\mumax{\sqrt{d}})^2}\mumax^4\right),$ (since $e^{-x}x\le 1, \forall x$).
\end{proof}

\begin{lemma}\label{seperated mu}
    For any $\GMM(\bmu)$, if there exists $k\in[n]$ such that $\|\mu_k-\mu_{\imax}\|\geq \frac{\mumax}{2}$, then we have
    \[\sum_{i,j\in[n]} \pi_i\pi_j\|\mu_i-\mu_j\|^2\geq \frac{\pimin}{8}\mumax^2.\]
\end{lemma}

\begin{proof}
By Cauchy–Schwarz inequality, we have $\|a\|^2+\|b\|^2\geq \frac{1}{2}\|a-b\|^2$, so for $\forall i\in[n]$ we have
\[\begin{split}
    \sum_{j\in[n]}\pi_j\|\mu_i-\mu_j\|^2&\geq \pi_{\imax}\|\mu_i-\mu_{\imax}\|^2 +\pi_{k}\|\mu_i-\mu_k\|^2\\
    &\geq  \frac{\pimin}{2}\|(\mu_i-\mu_{\imax})-(\mu_i-\mu_k)\|^2=\frac{\pimin}{2}\|\mu_k-\mu_{\imax}\|^2.
\end{split}\]

Therefore
    \begin{equation*}
        \begin{split}
            \sum_{i,j\in[n]} \pi_i\pi_j\|\mu_i-\mu_j\|^2=\sum_{i\in[n]}\pi_i\sum_{j\in[n]}\pi_j\|\mu_i-\mu_j\|^2\geq \sum_{i\in[n]}\pi_i\frac{\pimin}{2}\|\mu_k-\mu_{\imax}\|^2\geq \frac{\pimin}{8}\mumax^2,
        \end{split}
    \end{equation*}
    where the last inequality is because $\|\mu_k-\mu_{\imax}\|\geq \frac{\mumax}{2}$ and $\sum_i \pi_i=1$.
\end{proof}

\begin{lemma}\label{almost parallel mu}
    For any $\GMM(\bmu)$, if for $\forall k\in[n]$ we have $\|\mu_{\imax}-\mu_k\|< \frac{\mumax}{2}$, then
    \[\E_x\left[\|\tpsi_{\bmu}(x)\|^2\right]\geq\frac{1}{4}\mumax^2.\]
\end{lemma}

\begin{proof}
    For any $k\in[n]$, by Cauchy–Schwarz inequality we have 
    \begin{equation}\label{NL6}
        \begin{split}
        \langle \mu_k, \mu_{\imax}\rangle&=\langle \mu_{\imax}-(\mu_{\imax}-\mu_k), \mu_{\imax}\rangle=\|\mu_{\imax}\|^2-\ip{\mu_{\imax}-\mu_k,\mu_{\imax}}\\
        &\geq \mumax^2-\|\mu_{\imax}-\mu_k\|\mumax>\frac{1}{2}\mumax^2,
    \end{split}
    \end{equation}
    where the last inequality is because $\|\mu_{\imax}-\mu_k\|< \frac{\mumax}{2}$. 

    Note that \eqref{NL6} implies $ \langle \mu_k, \overline{\mu_{\imax}}\rangle>\frac{1}{2}\mumax$, so for $\forall x\in\R^d$ we have
    \begin{equation}\label{NL7}
        \|\tpsi_{\bmu}(x)\|=\left\|\sum_{k\in[n]}\psi_k(x)\mu_k\right\|\geq \ip{\sum_{k\in[n]}\psi_k(x)\mu_k, \overline{\mu_{\imax}}}=\sum_{k\in[n]}\psi_k(x)\ip{\mu_k, \overline{\mu_{\imax}}}>\frac{1}{2}\mumax,
    \end{equation}
    where we used $\sum_{k\in[n]}\psi_k(x)=1$ at the last inequality.
\end{proof}

\begin{replemma}{separation lower bound}
    For any $\GMM(\bmu)$ we have
    \[ \E_x\left[\|\tpsi_{\bmu}(x)\|^2\right]\geq \frac{\exp\left(-8U\right)}{40000 d(1+2\mumax{\sqrt{d}})^2}\left(\sum_{i,j\in[n]} \pi_i\pi_j\|\mu_i-\mu_j\|^2\right)^2.\]
\end{replemma}

\begin{proof}
The key idea is to consider the gradient of $\tpsi_{\bmu}$, which can be calculated as
\begin{equation}\label{nabla_tilde_psi}
    \begin{split}
        \nabla_x \tpsi_{\bmu}(x)
        &=\sum_i\mu_i \left(\frac{\partial \psi_i(x)}{\partial x} \right)^{\top}\\
        &=\sum_i\psi_i(x)\mu_i\mu_i^{\top}-\sum_{i,j}\psi_i(x)\psi_j(x)\mu_i\mu_j^{\top}\\
        &=\sum_{i,j\in[n]}\psi_i(x)\psi_j(x)\mu_i\mu_i^{\top}-\sum_{i,j}\psi_i(x)\psi_j(x)\mu_i\mu_j^{\top}\\
        &=\sum_{i,j\in[n]}\psi_i(x)\psi_j(x)\mu_i(\mu_i-\mu_j)^{\top}\\
        &=\sum_{i,j\in[n]}\psi_i(x)\psi_j(x)\frac{1}{2}\left(\mu_i(\mu_i-\mu_j)^{\top}+\mu_j(\mu_j-\mu_i)^{\top}\right)\\
        &=\frac{1}{2}\sum_{i,j\in[n]} \psi_i(x)\psi_j(x)(\mu_i-\mu_j)(\mu_i-\mu_j)^{\top},
    \end{split}
\end{equation}
where we used \eqref{psi gradient} in the second identity.

By Cauchy-Schwarz inequality, we have $\|a\|^2+\|b\|^2\geq \frac{1}{2}\|a-b\|^2$, which implies
\begin{equation}\label{tpsi to gradient}
    \begin{split}
    \E_x\left[\|\tpsi_{\bmu}(x)\|^2\right]&= \frac{1}{2}\E_x\left[\|\tpsi_{\bmu}(x)\|^2+\|\tpsi_{\bmu}(-x)\|^2\right]\\
    &\geq \frac{1}{4}\E_x\left[\left\|\tpsi_{\bmu}(x)-\tpsi_{\bmu}(-x)\right\|^2\right]\\
    &\geq \frac{1}{4}\E_x\left[\left\langle\tpsi_{\bmu}(x)-\tpsi_{\bmu}(-x), \overline{x}\right\rangle^2\right]\\
    &=\frac{1}{4}\E_x\left[\left(\int_{t=-1}^{1} \frac{\partial}{\partial t}\langle\tpsi_{\bmu}(tx), \overline{x}\rangle\diff t\right)^2\right]\\
    &=\frac{1}{4}\E_x\left[\left(\int_{t=-1}^{1} {x}^{\top}\nabla \tpsi_{\bmu}(tx)\overline{x}\diff t\right)^2\right]\\
    &=\frac{1}{4}\E_x\left[\left(\int_{t=-1}^{1} \|x\|\cdot\overline{x}^{\top}\nabla \tpsi_{\bmu}(tx)\overline{x}\diff t\right)^2\right],
\end{split}
\end{equation}
where we used $ \frac{\partial}{\partial t}\tpsi_{\bmu}(tx)=\nabla \tpsi_{\bmu}(tx)x$ at the second to last identity. Careful readers might notice that the term $\left(\int_{t=-1}^{1} \|x\|\cdot\overline{x}^{\top}\nabla \tpsi_{\bmu}(tx)\overline{x}\diff t\right)^2$ is not well-defined when $x= 0$, but we can still calculate its expectation over the whole probability space since the integration is only singular on a zero-measure set.
   
For each $x\neq 0$, by \eqref{nabla_tilde_psi} we have
\[\overline{x}^{\top}\nabla \tpsi_{\bmu}(tx)\overline{x}=\frac{1}{2}\sum_{i,j\in[n]} \psi_i(tx)\psi_j(tx)\langle\mu_i-\mu_j, \overline{x}\rangle^2.\]

So
\begin{equation}\label{NL4}
    \begin{split}
    &\quad\E_x\left[\|\tpsi_{\bmu}(x)\|^2\right]\\
    &\geq\frac{1}{16}\E_x\left[\left(\int_{t=-1}^{1} \|x\|\sum_{i,j\in[n]} \psi_i(tx)\psi_j(tx)\langle\mu_i-\mu_j, \overline{x}\rangle^2 \diff t\right)^2\right]\\
    &=\frac{1}{16}\E_x\left[\left( \|x\|\sum_{i,j\in[n]} \langle\mu_i-\mu_j, \overline{x}\rangle^2\int_{t=-1}^{1}\psi_i(tx)\psi_j(tx) \diff t\right)^2\right]\\
    &\geq \frac{1}{16}\E_x\left[\left( \|x\|\sum_{i,j\in[n]} \langle\mu_i-\mu_j, \overline{x}\rangle^2
    \frac{1}{2\mumax\|x\|}\pi_i\pi_j\exp\left(-4U\right)\left(1-\exp\left(-4\mumax\|x\|\right)\right)\right)^2\right]\\
    &=\frac{\exp\left(-8U\right)}{64}\E_x\left[\left( \sum_{i,j\in[n]} \pi_i\pi_j\langle\mu_i-\mu_j, \overline{x}\rangle^2
    \frac{1-\exp\left(-4\mumax\|x\|\right)}{\mumax}\right)^2\right]\\
    &\geq\frac{\exp\left(-8U\right)}{64}\left(\sum_{i,j\in[n]} \pi_i\pi_j\E_x\left[ \langle\mu_i-\mu_j, \overline{x}\rangle^2
    \frac{1-\exp\left(-4\mumax\|x\|\right)}{\mumax}\right]\right)^2
 \end{split}
\end{equation}
where we used Lemma \ref{integral psi(tx) bound lemma} at the fourth line and Cauchy-Schwarz inequality at the last line.

The last step is to lower bound $\E_x\left[ \langle\mu_i-\mu_j, \overline{x}\rangle^2
    \left(1-\exp\left(-4\mumax\|x\|\right)\right)/\mumax\right]$. Since $x$ is sampled from $\N(0,I_d)$, which is spherically symmetric, we know that the two random variables $\{\overline{x}, \|x\|\}$ are independent. Therefore
    \begin{equation}\label{NL1}
        \E_x\left[ \langle\mu_i-\mu_j, \overline{x}\rangle^2
    \frac{1-\exp\left(-4\mumax\|x\|\right)}{\mumax}\right]=\E_x\left[ \langle\mu_i-\mu_j, \overline{x}\rangle^2\right]
    \E_x\left[\frac{1-\exp\left(-4\mumax\|x\|\right)}{\mumax}\right].
    \end{equation}
    For the first term in \eqref{NL1}, we have $\E_x\left[ \langle\mu_i-\mu_j, \overline{x}\rangle^2\right]=\|\mu_i-\mu_j\|^2/d$ since $\overline{x}$ is spherically symmetrically distributed. By norm-concentration inequality of Gaussian \citep{dasgupta2013two} we know that $\Pr\left[\|x\|\geq \frac{\sqrt{d}}{2}\right]\geq 1/50, \forall d$. The second term in \eqref{NL1}  can be therefore lower bounded as
    \begin{equation}\label{NL2}
        \begin{split}
            \E_x\left[\frac{1-\exp\left(-4\mumax\|x\|\right)}{\mumax}\right] \geq \Pr\left[\|x\|\geq \frac{\sqrt{d}}{2}\right]\frac{1-\exp\left(-4\mumax\cdot\frac{\sqrt{d}}{2}\right)}{\mumax}\geq \frac{1-\exp\left(-2\mumax{\sqrt{d}}\right)}{50\mumax}.
        \end{split}
    \end{equation}

    Plugging \eqref{NL2} into \eqref{NL1}, we get
    \begin{equation}\label{NL3}
        \E_x\left[ \langle\mu_i-\mu_j, \overline{x}\rangle^2
    \frac{1-\exp\left(-4\mumax\|x\|\right)}{\mumax}\right]\geq
    \frac{1-\exp\left(-2\mumax{\sqrt{d}}\right)}{50d\mumax}\|\mu_i-\mu_j\|^2.
    \end{equation}

    Now we can plug \eqref{NL3} into \eqref{NL4} and get

    \begin{equation}\label{NL5}
    \begin{split}
    \E_x\left[\|\tpsi_{\bmu}(x)\|^2\right]
    &\geq\frac{\exp\left(-8U\right)}{64}\left(\sum_{i,j\in[n]} \pi_i\pi_j\E_x\left[ \langle\mu_i-\mu_j, \overline{x}\rangle^2
    \frac{1-\exp\left(-4\mumax\|x\|\right)}{\mumax}\right]\right)^2\\
    &\geq \frac{\exp\left(-8U\right)}{64}\left(\sum_{i,j\in[n]} \pi_i\pi_j\frac{1-\exp\left(-2\mumax{\sqrt{d}}\right)}{50d\mumax}\|\mu_i-\mu_j\|^2\right)^2\\
    &\geq \frac{\exp\left(-8U\right)}{64}\left(\sum_{i,j\in[n]} \pi_i\pi_j\frac{1-\frac{1}{1+2\mumax{\sqrt{d}}}}{50d\mumax}\|\mu_i-\mu_j\|^2\right)^2\\
    &=\frac{\exp\left(-8U\right)}{40000 d(1+2\mumax{\sqrt{d}})^2}\left(\sum_{i,j\in[n]} \pi_i\pi_j\|\mu_i-\mu_j\|^2\right)^2
 \end{split}
\end{equation}
 where we used the inequality $\forall t\geq 0, e^{-t}\leq \frac{1}{1+t}$ at the second to last line.
\end{proof}

\begin{reptheorem}{Global convergence of gradient EM}
Consider training a student $n$-component GMM initialized from $\bm{\mu}(0) = (\mu_1(0)^{\top},\ldots, \mu_n(0)^{\top})^{\top}$ to learn a single-component ground truth GMM $\N(0, I_d)$ with population gradient EM algorithm. If the step size satisfies $\eta \leq O\left(\frac{\exp\left(-8U(0)\right)\pimin^2}{n^2d^2(\frac{1}{\mumax(0)}+\mumax(0))^2}\right)$, then gradient EM converges globally with rate
\[\loss(\bmu(t))\leq \frac{1}{\sqrt{\gamma t}},\]
where  $\gamma = \Omega\left(\frac{\eta\exp\left(-16U(0)\right)\pimin^4}{n^2d^2(1+\mumax(0){\sqrt{dn}})^4}\right)\in\R^+$. Recall that $\mumax(0)=\max\{\|\mu_1(0)\|,\ldots, \|\mu_n(0)\|\}$ and $U(0)=\sum_{i\in[n]}\|\mu_i(0)\|^2$ are two initialization constants.
\end{reptheorem}

\begin{proof}
We use mathematical induction to prove Theorem \ref{Global convergence of gradient EM}, by proving the following two conditions inductively:
    \begin{equation}\label{mu norm upper bound}
         U(t)\leq U(0)=\sum_{i\in [n]}\|\mu_i(0)\|^2, \forall t.
    \end{equation}

    \begin{equation}\label{convergence of loss}
        \frac{1}{\loss^2(\bmu(t))}\geq \gamma t+\frac{1}{\loss^2(\bmu(0))} ,\forall t.
    \end{equation}
Note that \eqref{convergence of loss} directly implies the theorem, so now we just need to prove \eqref{mu norm upper bound} and \eqref{convergence of loss} together.

The induction base for $t=0$ is trivial. Now suppose the conditions hold for time step $t$, consider $t+1$. By induction hypothesis \eqref{mu norm upper bound} we have $\|\mu_i(t)\|\leq\mumax(t)\leq \sqrt{n}\mumax(0), \forall t$.
 
\bftext{Proof of \eqref{convergence of loss}.}
Since $\nabla_{\bmu} Q(\bmu|\bmu) = \nabla_{\bmu} \loss(\bmu)$, we can apply classical analysis of gradient descent \citep{nesterov2018lectures} as
\begin{equation}\label{NL18}
    \begin{split}
        &\quad \loss(\bmu(t+1))-\loss(\bmu(t)) \\
        &= \loss(\bmu(t) - \eta\nabla \loss(\bmu(t)) )-\loss(\bmu(t)) \\
        &=-\int_{s=0}^1 \ip{\nabla\loss(\bmu(t) - s\eta\nabla \loss(\bmu(t)) ), \eta\nabla \loss(\bmu(t))} \diff s\\
        &= -\int_{s=0}^1 \ip{\nabla\loss(\bmu(t)), \eta\nabla \loss(\bmu(t))} \diff s 
        + \int_{s=0}^1 \ip{\nabla\loss(\bmu(t))-\nabla\loss(\bmu(t) - s\eta\nabla \loss(\bmu(t)) ), \eta\nabla \loss(\bmu(t))} \diff s\\
        &= -\eta\|\nabla\loss(\bmu(t))\|^2 + \eta \int_{s=0}^1\ip{ \nabla\loss(\bmu(t))-\nabla\loss(\bmu(t) - s\eta\nabla \loss(\bmu(t)) ), \nabla \loss(\bmu(t))}\diff s \\
    \end{split}
\end{equation}
Note that the gradient norm can be upper bounded as
\[\begin{split}
\|\nabla_{\mu_i}\loss(\bmu(t))\|&=\left\|\E_x\left[\psi_i(x)\sum_{k\in[n]}\psi_k(x)\mu_k(t)\right]\right\|\leq\E_x\left[\psi_i(x)\sum_{k\in[n]}\psi_k(x)\left\|\mu_k(t)\right\|\right]\\
&\leq \sum_{k}\|\mu_k(t)\|\leq \sqrt{nU(t)}\leq n\mumax(0).
\end{split}\]
Then for any $s\in[0,1]$, we have $\|s\eta\nabla_{\mu_i} \loss(\bmu(t))\|\leq \eta n\mumax(0)\leq \frac{1}{\max\left\{6d,2\|\mu_i(t)\|\right\}}$.
So we can apply Theorem \ref{smoothness} and get
 \[\begin{split}
     &\quad\|\nabla_{\mu_i}\loss(\bmu(t))-\nabla_{\mu_i}\loss(\bmu(t) - s\eta\nabla_{\mu_i} \loss(\bmu(t)) )\|\\
     &\leq n\mumax(t)(30\sqrt{d}+4\mumax(t))\| s\eta\nabla_{\mu_i} \loss(\bmu(t))\|+\sum_{k\in[n]}\|s\eta\nabla_{\mu_k} \loss(\bmu(t))\|.
 \end{split}\]

Therefore for $\forall s\in[0,1]$,
\begin{equation}\label{NL17}
    \begin{split}
        &\quad \ip{ \nabla\loss(\bmu(t))-\nabla\loss(\bmu(t) - s\eta\nabla \loss(\bmu(t)) ), \nabla \loss(\bmu(t))}\\
        &\leq \sum_{i\in[n]} \|\nabla_{\mu_i}\loss(\bmu(t))-\nabla_{\mu_i}\loss(\bmu(t) - s\eta\nabla_{\mu_i} \loss(\bmu(t)) )\|\cdot \|\nabla_{\mu_i}\loss(\bmu(t))\|\\
        &\leq \sum_{i\in[n]} \left(n\mumax(t)(30\sqrt{d}+4\mumax(t))\| s\eta\nabla_{\mu_i} \loss(\bmu(t))\|+\sum_{k\in[n]}\|s\eta\nabla_{\mu_k} \loss(\bmu(t))\|\right) \|\nabla_{\mu_i}\loss(\bmu(t))\|\\
        &\leq \eta\left(n\mumax(t)(30\sqrt{d}+4\mumax(t))+n^2\right)\|\nabla\loss(\bmu(t))\|^2\\
         &\leq \eta\left(4n^2\maxz^2+30\sqrt{d}n^{3/2}\maxz+n^2\right)\|\nabla\loss(\bmu(t))\|^2\\
         &\leq 20\eta\sqrt{d}n^2(\mumax^2(0)+1)\|\nabla\loss(\bmu(t))\|^2.\\
    \end{split}
\end{equation}

Plugging \eqref{NL17} into \eqref{NL18}, since $\eta\leq O\left(\frac{1}{\sqrt{d}n^2(\mumax^2(0)+1)}\right)$ we have
\begin{equation}\label{NL20}
    \loss(\bmu(t+1))-\loss(\bmu(t)) \leq -\eta\|\nabla\loss(\bmu(t))\|^2 +20\eta\sqrt{d}n^2(\mumax^2(0)+1)\|\nabla\loss(\bmu(t))\|^2\leq -\frac{\eta}{2}\|\nabla\loss(\bmu(t))\|^2 .
\end{equation}

By Lemma \ref{Gradient projection lower bound} we can lower bound the gradient norm as
\begin{equation}\label{NL19}
    \begin{split}
    &\|\nabla\loss(\bmu(t))\|\geq \frac{\ip{\nabla\loss(\bmu(t)), \bmu(t)}}{\|\bmu(t)\|}\geq \frac{\ip{\nabla\loss(\bmu(t)), \bmu(t)}}{n\mumax(t)}\geq
\Omega\left(\frac{\exp\left(-8U(t)\right)\pimin^2}{nd(1+\mumax(t){\sqrt{d}})^2}\right)\mumax^3(t)\\
&\overset{\text{Theorem \ref{Loss function upper bound}}}{\geq} \Omega\left(\frac{\exp\left(-8U(t)\right)\pimin^2}{nd(1+\mumax(t){\sqrt{d}})^2}\right)(2\loss(\bmu(t))^{3/2}
\geq\Omega\left(\frac{\exp\left(-8U(0)\right)\pimin^2}{nd(1+\mumax(0){\sqrt{dn}})^2}\right)\loss^{3/2}(\bmu(t)).
\end{split}
\end{equation}
  Combining \eqref{NL19} and \eqref{NL20}, we have
  \begin{equation}\label{NL21}
      \loss(\bmu(t+1))\leq \loss(\bmu(t)) - \frac{\eta}{2}\|\nabla\loss(\bmu(t))\|^2\leq \loss(\bmu(t)) -\Omega\left(\frac{\eta\exp\left(-16U(0)\right)\pimin^4}{n^2d^2(1+\mumax(0){\sqrt{dn}})^4}\right)\loss^{3}(\bmu(t)).
  \end{equation}

Note that the above inequality implies  $\loss(\bmu(t+1))\leq \loss(\bmu(t))$, therefore
\[\begin{split}
    &\quad\frac{1}{\loss^2(\bmu(t+1))}-\frac{1}{\loss^2(\bmu(t))}=\frac{(\loss(\bmu(t))-\loss(\bmu(t+1)))(\loss(\bmu(t))+\loss(\bmu(t+1)))}{\loss^2(\bmu(t))\loss^2(\bmu(t+1))}\\
    &\geq \frac{(\loss(\bmu(t))-\loss(\bmu(t+1))\loss(\bmu(t))}{\loss^4(\bmu(t))}
    \overset{\eqref{NL21}}{\geq}\Omega\left(\frac{\eta\exp\left(-16U(0)\right)\pimin^4}{n^2d^2(1+\mumax(0){\sqrt{dn}})^4}\right)=\gamma.
\end{split} \]

On the other hand, by induction hypothesis we have $\frac{1}{\loss^2(\bmu(t))}\geq \gamma t+\frac{1}{\loss^2(\bmu(0))}$, combined with the above inequality, we have $\frac{1}{\loss^2(\bmu(t+1))}\geq \frac{1}{\loss^2(\bmu(t))}+\gamma\geq \gamma (t+1)+\frac{1}{\loss^2(\bmu(0))}$, which finishes the proof of \eqref{convergence of loss}.
 
\bftext{Proof of \eqref{mu norm upper bound}.}
The dynamics of potential function $U$ can be calculated as
\begin{equation}\label{NL23}
        \begin{split}
            &\quad U(\bmu(t+1))= \sum_{i\in[n]}\left\|\mu_i(t+1)\right\|^2\\
            &=\sum_{i\in[n]}\left\|\mu_i(t)-\eta\nabla_{\mu_i} Q(\bmu(t)|\bmu(t))\right\|^2\\
            &=U(\bmu(t))
            -{\eta\sum_{i\in[n]}\ip{\mu_i(t), \nabla_{\mu_i} Q(\bmu(t)|\bmu(t))}}
            +{\eta^2\sum_{i\in[n]}\|\nabla_{\mu_i} Q(\bmu(t)|\bmu(t))\|^2}\\
            &\overset{\text{Corollary \ref{Gradient projection lemma}}}{=}U(\bmu(t))
            -\underbrace{\eta\E_x\left[\|\tpsi_{\bmu(t)}(x)\|^2\right]}_{I_1}
            +\underbrace{\eta^2\sum_{i\in[n]}\|\nabla_{\mu_i} Q(\bmu(t)|\bmu(t))\|^2}_{I_2}.
        \end{split}
\end{equation}

By induction hypothesis, the first term $I_1$ can be bounded by Lemma \ref{Gradient projection lower bound} as 
\begin{equation}\label{NL22}
    I_1\geq \eta \Omega\left(\frac{\exp\left(-8U(t)\right)\pimin^2}{d(1+\mumax(t){\sqrt{d}})^2}\right)\mumax^4(t)\geq \eta \Omega\left(\frac{\exp\left(-8U(0)\right)\pimin^2}{n^2d(1+\mumax(0){\sqrt{nd}})^2}\right)U^2(\bmu(t)).
\end{equation}

The second term $I_2$ is a perturbation term that can be upper bounded by Lemma \ref{Gradient transformation lemma} as
\begin{equation}\label{NL8}
    \begin{split}
    I_2&={\eta^2\sum_{i\in[n]}\|\nabla_{\mu_i} Q(\bmu(t)|\bmu(t))\|^2}=\eta^2\sum_{i\in[n]}\left\|\E_x\left[\psi_i(x)\sum_{k\in[n]}\psi_k(x)\mu_k(t)\right]\right\|^2\\
    &\leq \eta^2\sum_{i\in[n]}\E_x\left[\left\|\psi_i(x)\sum_{k\in[n]}\psi_k(x)\mu_k(t)\right\|\right]^2\\
    &\leq \eta^2\sum_{i\in[n]}\E_x\left[\psi_i(x)\sum_{k\in[n]}\psi_k(x)\left\|\mu_k(t)\right\|\right]^2\\
    &\leq \eta^2\sum_{i\in[n]}\E_x\left[\sqrt{\left(\sum_{k\in[n]}\psi_i^2(x)\psi_k^2(x)\right)\left(\sum_{k\in[n]}\|\mu_k(t)\|^2\right)}\right]^2\\
    &\leq \eta^2\sum_{i\in[n]}\E_x\left[\sum_{k\in[n]}\psi_i^2(x)\psi_k^2(x)\right]\E_x\left[\sum_{k\in[n]}\|\mu_k(t)\|^2\right]\\
    &=\eta^2 U(\bmu(t))\E_x\left[\sum_{i\in[n]}\sum_{k\in[n]}\psi_i^2(x)\psi_k^2(x)\right]\\
    &\leq \eta^2 U(\bmu(t))\E_x\left[\left(\sum_{i\in[n]}\psi_i(x)\right)\left(\sum_{k\in[n]}\psi_k(x)\right)\right]\\
    &=\eta^2 U(\bmu(t)).
\end{split}
\end{equation}
where we use triangle inequality twice at the second and third line, and Cauchy-Schwarz inequality twice at the fourth and fifth line.

Putting \eqref{NL8}, \eqref{NL22} and \eqref{NL23} together, we get

\[U(\bmu(t+1))\leq U(\bmu(t))-\eta \Omega\left(\frac{\exp\left(-8U(0)\right)\pimin^2}{n^2d(1+\mumax(0){\sqrt{nd}})^2}\right)U^2(\bmu(t))+\eta^2U(\bmu(t)).\]

Consider two cases:

a). If $\frac{U(0)}{2} \leq U(\bmu(t))\leq U(0) $, then
    \[\begin{split}
        &\quad U(\bmu(t+1))\leq U(\bmu(t))-\eta U(\bmu(t))\left(\Omega\left(\frac{\exp\left(-8U(0)\right)\pimin^2}{n^2d(1+\mumax(0){\sqrt{nd}})^2}\right)U(\bmu(t))-\eta\right)\\
        &\leq U(\bmu(t))-\eta U(\bmu(t))\left(\Omega\left(\frac{\exp\left(-8U(0)\right)\pimin^2}{n^2d(1+\mumax(0){\sqrt{nd}})^2}\right)\frac{n}{2}\mumax^2(0)-\eta\right)\leq U(\bmu(t))\leq n\mumax^2(0),
    \end{split}\]
note that we used $\eta \leq O\left(\frac{\exp\left(-8U(0)\right)\pimin^2}{n^2d(1+\mumax(0){\sqrt{nd}})^2}\right)\frac{n}{2}\mumax^2(0)$.

b). If $ U(\bmu(t))<\frac{1}{2}U(0)$, then $U(\bmu(t+1))\leq (1+\eta^2)U(\bmu(t))\leq 2U(\bmu(t))\leq U(0)$.

Since \eqref{mu norm upper bound} holds in both cases, our proof is done.
\end{proof}

\subsection{Proofs for Section \ref{Section counter-examples}}
\label{Proofs for counter-example theorem}

\begin{replemma}{Gradient norm upper bound}
For any $\bmu$ satisfying $\|\mu_1\|,\|\mu_2\|\geq 10\sqrt{d}, \|\mu_3\|,\ldots,\|\mu_n\|\leq \sqrt{d}$, the gradient of $\loss$ at $\bmu$ can be upper bounded as
    \[\|\nabla_{\mu_i}\loss(\bmu)\| \leq 2(\|\mu_3\|+\cdots+\|\mu_n\|)+2\exp(-d)(\|\mu_1\|+\|\mu_2\|),\forall i\in[n].\]
\end{replemma}
\begin{proof}
    Recall that the gradient has the form $\nabla_{\mu_i}\loss(\bmu)=\E_x\left[\psi_i(x)\sum_{k\in[n]}\psi_k(x)\mu_k\right],$ hence its norm can be upper bounded as

    \begin{equation}\label{NL25}
        \begin{split} &\quad\|\nabla_{\mu_i}\loss(\bmu)\|\leq\E_x\left[\psi_i(x)\sum_{k\in[n]}\psi_k(x)\|\mu_k\|\right]\\
    &\leq \E_x\left[\sum_{k\in[n]}\psi_k(x)\|\mu_k\|\Bigg\vert \|x\|\leq 2\sqrt{d}\right]+\E_x\left[\sum_{k\in[n]}\psi_k(x)\|\mu_k\|\Bigg\vert \|x\|> 2\sqrt{d}\right]\Pr\left[\|x\|> 2\sqrt{d}\right].
    \end{split}
    \end{equation}

    For any $ \|x\|\leq 2\sqrt{d}$ and $i>2$, we have $\exp(-\|x-\mu_i\|^2/2)\geq \exp(-(\|x\|+\|\mu_i\|)^2/2)\geq \exp(-9d/2)$, while for $i\in\{1,2\}$, $\exp(-\|x-\mu_i\|^2/2)\leq \exp(-(\|\mu_i\|-\|x\|)^2/2)\leq \exp(-(10\sqrt{d}-2\sqrt{d})^2/2)=\exp(-32d)$. Since $\psi_i(x)\propto\exp(-\|x-\mu_i\|^2/2)$ we have
    \[\|x\|\leq 2\sqrt{d} \To \psi_i(x)\leq \frac{\exp(-\|x-\mu_i\|^2/2)}{\exp(-\|x-\mu_1\|^2/2)}\leq \frac{\exp(-32d)}{\exp(-9d/2)}\leq \exp(-25d), \forall i\in\{1,2\}. \]

    Therefore the first term in \eqref{NL23} can be bounded as $\E_x\left[\sum_{k\in[n]}\psi_k(x)\|\mu_k\|\Bigg\vert \|x\|\leq 2\sqrt{d}\right]\leq (\|\mu_3\|+\cdots+\|\mu_n\|)+\exp(-25d)(\|\mu_1\|+\|\mu_2\|).$

    On the other hand, by tail bound of the norm of Gaussian vectors (see Lemma 8 of \citep{yan_convergence_2017}) we have $\Pr\left[\|x\|> 2\sqrt{d}\right]\leq \exp(-d)$. Putting everything together, \eqref{NL25} can be further bounded as
    \begin{equation*}\label{NL24}
        \begin{split} \|\nabla_{\mu_i}\loss(\bmu)\|&\leq(\|\mu_3\|+\cdots+\|\mu_n\|)+\exp(-25d)(\|\mu_1\|+\|\mu_2\|)+\exp(-d)\sum_{i\in[n]}\|\mu_i\|\\
        &\leq 2(\|\mu_3\|+\cdots+\|\mu_n\|)+2\exp(-d)(\|\mu_1\|+\|\mu_2\|).
    \end{split}
    \end{equation*}
\end{proof}

\begin{reptheorem}{Existence of bad initialization region}
For any $n\geq 3$, define $\tilde{\bmu}(0)=({\mu}^{\top}_1(0),\ldots,{\mu}^{\top}_n(0))$ as follows: $ {\mu}_1(0)=12\sqrt{d}e_1, {\mu}_2(0)=-12\sqrt{d}e_1, \mu_3(0)=\cdots=\mu_n(0)=0$, where $e_1$ is a standard unit vector. Then population gradient EM initialized with means $\tilde{\bmu}(0)$ and equal weights $\pi_1=\ldots=\pi_n=1/n$ will be trapped in a bad local region around $\tilde{\bmu}(0)$ for exponentially long time $T=\frac{1}{30\eta}e^{d}=\frac{1}{30\eta}\exp(\Theta(U(0)))$. More rigorously, for any $0\leq t\leq T, \exists i\in[n]$ such that
    \[ \|\mu_i(t)\|\geq 10\sqrt{d}, \;\; \]
\end{reptheorem}

\begin{proof}

We prove the following statement inductively: %for 
$\forall\; 0\leq t\leq T$:
\begin{equation}\label{symmetry condition}
   {\mu}_1(t)+{\mu}_2(t)=0, \mu_3(t)=\cdots=\mu_n(t)=0
\end{equation}
\begin{equation}\label{Bad local trap}
    \forall i,\; \|\mu_i(t)-\mu_i(0)\|\leq \eta t (60\sqrt{d}e^{-d}).
\end{equation}

\eqref{symmetry condition} states that during the gradient EM update, $\mu_1$ will keep stationary at $0$. while the symmetry between $\mu_2,\ldots,\mu_n$ will be preserved.

The induction base is trivial. Now suppose \eqref{Bad local trap}, \eqref{symmetry condition} holds for $0,1,\ldots, t$, we prove the case for $t+1$.

\bftext{Proof of \eqref{symmetry condition}. }
Due to the induction hypothesis, one can see from direct calculation that $ \forall x, $ we have $\psi_i(x|\bmu(t)) = \psi_i(-x|\bmu(t))$ for $i=3,\ldots, n$, and $ \psi_1(x|\bmu(t))=\psi_{2}(-x|\bmu(t))$.

Consequently for $\forall i>2$ we have
\[\begin{split}
&\quad \nabla_{\mu_i}\loss(\bmu(t))=\E_x\left[\psi_i(x|\bmu(t))\sum_{k\in[n]}\psi_k(x|\bmu(t))\mu_k(t)\right]=\E_x\left[\psi_i(x)(\psi_{1}(x)\mu_1(t)+\psi_{2}(x)\mu_2(t))\right]\\
&=\frac{1}{2}\E_x\left[\psi_i(x)(\psi_{1}(x)\mu_1(t)+\psi_{2}(x)\mu_2(t))+\psi_i(-x)(\psi_{1}(-x)\mu_1(t)+\psi_{2}(-x)\mu_2(t))\right]\\
&=\frac{1}{2}\E_x\left[\psi_i(x)(\psi_1(x)(\mu_1(t)+\mu_2(t))+\psi_{2}(x)(\mu_2(t)+\mu_1(t)))\right]=0 \To \mu_1it+1)=\mu_i(t)=0.
\end{split}\]
Similarly, for $\mu_1,\mu_2$ we have
\[\begin{split}
&\quad \nabla_{\mu_1}\loss(\bmu(t))=\E_x\left[\psi_1(x|\bmu(t))\sum_{k\in[n]}\psi_k(x|\bmu(t))\mu_k(t)\right]=\E_x\left[\psi_1(x)(\psi_{1}(x)\mu_1+\psi_{2}(x)\mu_{2})\right]\\
&=\E_x\left[\psi_2(-x)(\psi_{2}(-x)\mu_1+\psi_{1}(-x)\mu_{2})\right]=-\E_x\left[\psi_2(-x)(\psi_{2}(-x)\mu_2+\psi_{1}(-x)\mu_{1})\right]=-\nabla_{\mu_2}\loss(\bmu(t)).
\end{split}\]

This combined with the induction hypothesis implies $\mu_2(t+1)=-\mu_1(t+1)$, \eqref{symmetry condition} is proved.

\bftext{Proof of \eqref{Bad local trap}. }

By induction hypothesis, we have $\forall i,\; \|\mu_i(t)-\mu_i(0)\|\leq \eta t\cdot(60\sqrt{d}e^{-d})\leq \eta T\cdot(60\sqrt{d}e^{-d})\leq 2\sqrt{d}.$ So $\forall i\in\{1,2\}, \|\mu_i(t)\|\leq 
\|\mu_i(0)\|+2\sqrt{d}<15\sqrt{d}$. Then by Lemma \ref{Gradient norm upper bound}, $\forall i\in[n]$ we have
\[\|\nabla_{\mu_i}\loss(\bmu(t))\| \leq 2(\|\mu_3\|+\cdots+\|\mu_n\|)+2\exp(-d)(\|\mu_1\|+\|\mu_2\|)\leq 4\exp(-d)\cdot15\sqrt{d}=60\sqrt{d}e^{-d},\]
note that here we used $\mu_3(t)=\cdots=\mu_n(t)=0$. Therefore by the induction hypothesis we have $\|\mu_i(t+1)-\mu_i(0)\|\leq \eta t\cdot(60\sqrt{d}e^{-d})+\eta\|\nabla_{\mu_i}\loss(\bmu(t))\|\leq \eta (t+1)\cdot(60\sqrt{d}e^{-d})$, \eqref{Bad local trap} is proven.

By \eqref{Bad local trap}, $\forall 0\leq t\leq T$, for $i=1,2$ we have $\|\mu_i(t)\|\geq \|\mu_i(0)\|-\|\mu_i(t)-\mu_i(0)\|\geq 12\sqrt{d}-\eta T(60\sqrt{d}e^{-d})\geq 12\sqrt{d}-2\sqrt{d}=10\sqrt{d}.$ Our proof is done.
\end{proof}

\end{document}